\documentclass[acmtog]{acmart}

\acmSubmissionID{256}

\setcopyright{acmlicensed}
\acmJournal{TOG}
\acmYear{2023} \acmVolume{42} \acmNumber{6} \acmArticle{249} \acmMonth{12} \acmPrice{15.00}\acmDOI{10.1145/3618327}

\usepackage{color}
\usepackage{xcolor}
\usepackage{graphicx}
\usepackage{booktabs}
\usepackage{amsmath, amsthm, amsfonts, amssymb}
\usepackage{mathrsfs}
\usepackage{textcomp}
\usepackage{epstopdf}
\usepackage{multirow}
\usepackage{wrapfig}
\usepackage{subfig}
\usepackage{booktabs} 
\usepackage[ruled]{algorithm2e} 
\usepackage{algorithmicx}  
\usepackage{algpseudocode}
\usepackage{ragged2e}
\usepackage[normalem]{ulem}
\usepackage{cleveref}
\usepackage{bm}

\SetAlFnt{\small}
\SetAlCapFnt{\small}
\SetAlCapNameFnt{\small}
\SetAlCapHSkip{0pt}

\definecolor{green}{rgb}{0, 0.5, 0}
\definecolor{orange}{rgb}{0.6, 0.3, 0.1}
\definecolor{red}{rgb}{1.0, 0.0, 0.0}
\definecolor{teal}{rgb}{0.0, 0.4, 0.4}
\definecolor{purple}{rgb}{0.65,0,0.65}
\definecolor{saffron}{rgb}{0.95,0.75,0.2}
\definecolor{turquoise}{rgb}{0.0,0.5,0.5}
\definecolor{brown}{rgb}{0.5, 0.16, 0.16}
\definecolor{brickred}{rgb}{.6, .2 .1}
\definecolor{coral}{rgb}{1,0.45,0.33}
\definecolor{newcolor}{rgb}{.8,.349,.1}

\begin{document}

\title{Interaction-Driven Active 3D Reconstruction with Object Interiors}

\author{Zihao Yan}
\email{mr.salingo@gmail.com}
\affiliation{%
	\institution{Shenzhen University}
	\country{China}	
}

\author{Fubao Su}
\email{jyily.work@gmali.com}
\affiliation{%
	\institution{Shenzhen University}
	\country{China}	
}

\author{Mingyang Wang}
\email{michael.gonw@gmail.com}
\affiliation{%
	\institution{Shenzhen University}
	\country{China}	
}

\author{Ruizhen Hu}
\email{ruizhenhu@gmail.com}
\affiliation{%
	\institution{Shenzhen University}
	\country{China}	
}

\author{Hao Zhang}
\email{hao.r.zhang@gmail.com}
\affiliation{%
	\institution{Simon Fraser University}
	\country{Canada}	
}

\author{Hui Huang}
\email{hhzhiyan@gmail.com}
\authornote{Corresponding author: Hui Huang (hhzhiyan@gmail.com)}
\affiliation{%
	\institution{Shenzhen University}
	\department{College of Computer Science \& Software Engineering}
	\country{China}	
}

\renewcommand\shortauthors{Z. Yan, F. Su, M. Wang, R. Hu, H. Zhang, and H. Huang}

\begin{abstract}

We introduce an {\em active 3D reconstruction\/} method which integrates visual perception, {\em robot-object interaction\/}, and 3D
scanning to recover both the exterior and {\em interior\/}, i.e., unexposed, geometries of a target 3D object.
Unlike other works in active vision which focus on optimizing camera viewpoints to better investigate the environment, the primary 
feature of our reconstruction is an analysis of the {\em interactability\/} of various parts of the target object and the ensuing part manipulation by a robot to enable scanning of occluded regions. As a result, an understanding of part articulations of the
target object is obtained on top of complete geometry acquisition. 
Our method operates {\em fully automatically\/} by a Fetch robot with built-in RGBD sensors. It iterates between interaction analysis 
and interaction-driven reconstruction, scanning and reconstructing detected moveable parts one at a time, where both the articulated
part detection and mesh reconstruction are carried out by neural networks. In the final step, all the remaining, non-articulated parts, 
including all the interior structures that had been exposed by prior part manipulations and subsequently scanned, are reconstructed 
to complete the acquisition.
We demonstrate the performance of our method via qualitative and quantitative evaluation, ablation studies, comparisons to alternatives, as well as experiments in a real environment.

\end{abstract}



\begin{teaserfigure}
  \centering
  \includegraphics[width=\linewidth]{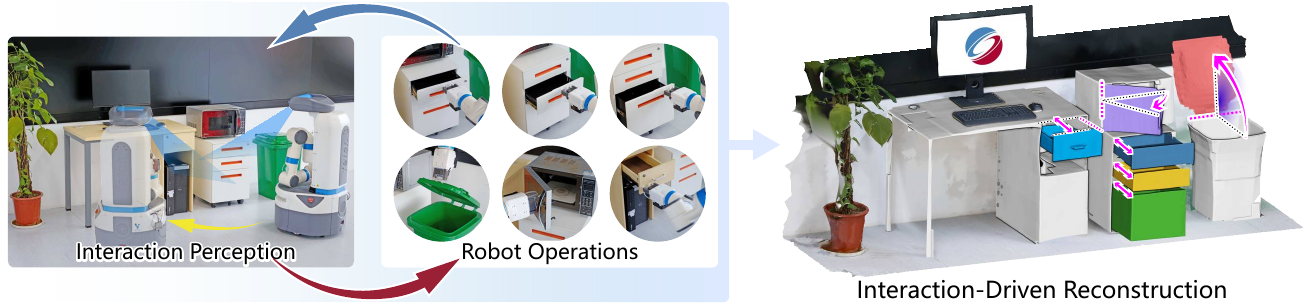}
  \caption{We introduce a {\em fully automatic\/}, {\em active 3D reconstruction\/} method which integrates interaction perception from depth sensors, real robot-object interaction (e.g., opening drawers), and on-the-fly scanning and reconstruction to obtain a {\em complete\/} geometry acquisition of both the object exteriors {\em and interiors}.}
  \Description{}
  \label{fig:teaser}
\end{teaserfigure}

\maketitle

\section{Introduction}
\label{sec:intro}

3D shape reconstruction has been one of the most classical problems in computer graphics. To date, the dominant majority of the proposed methods, whether geometry-based~\cite{berger2017survey} or learning-based~\cite{NF_survey}, can only capture the {\em exterior\/} surfaces of the target 3D object, since only these surfaces are exposed to the acquisition devices, such as a camera or a laser scanner. 
Such models can be well visualized but not properly {\em interacted with\/}, especially when the interactions can lead to part articulations that may reveal the objects' inside or {\em interiors\/}. For example, when a cabinet door or desk drawer is opened, the interior structures of these furniture items will be exposed, but they may have never been acquired to start with.
In interactive applications such as VR/AR, games, smart homes, and embodied AI, it is highly desirable to endow the 3D models therein with not only interior structures but also part-level motion attributes, so as to present a realistic user experience.

In this paper, we introduce an {\em active 3D reconstruction\/} method which integrates visual perception, {\em robot-object interaction\/}, and 3D scanning to recover both the exterior and interior geometries of a target 3D object. In our problem setting, the object interiors refer to
{\em unexposed} structures of the object relative to its {\em initial state\/} or articulated pose when it is being scanned. Such interiors can only be captured by applying proper motions to one or more of the object's parts to reveal them.
Our approach falls into the general realm of active perception~\cite{gibson1966, bajcsy1988active}, which seeks to understand the world around us by moving around and exploring it. However, unlike other works in active vision~\cite{aloimonos1998active} which focus on optimizing camera viewpoints to better investigate the environment, the primary feature of our reconstruction is an iterative analysis of the {\em interactability\/} of various parts of the target object and the ensuing part manipulation by a robot, e.g., opening the drawer of a cabinet, to reveal the unexposed geometries and scan both the outside and inside of the object. As a result, our method not only achieves complete geometry acquisition but also an understanding of part articulations.

\begin{figure*}[!t]
\centering
\includegraphics[width=\textwidth]{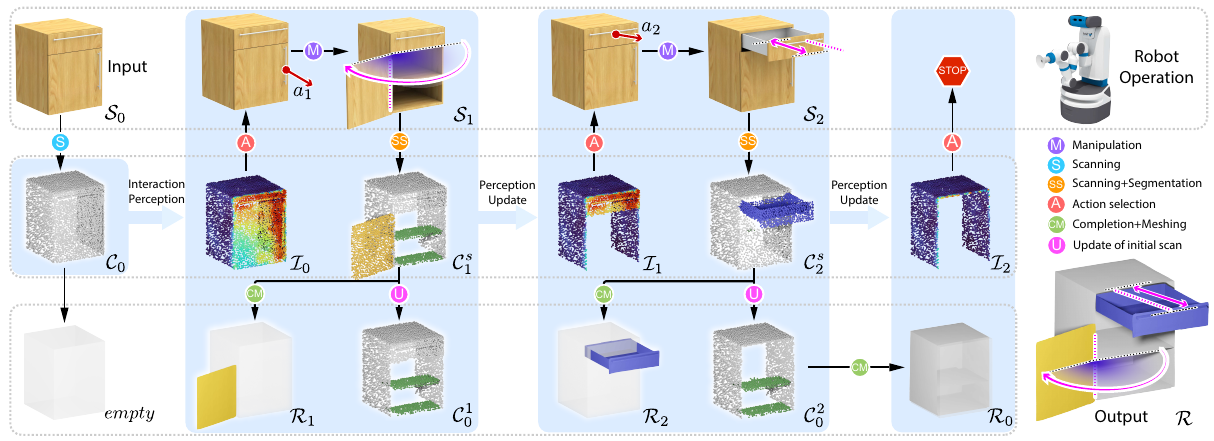}
\caption{A pipeline overview and running example for our active, interaction-driven 3D reconstruction method. Given a target 3D object in its rest state $\mathcal{S}_0$, our robot first obtains a scan $\mathcal{C}_0$ and infers its associated interactability map $\mathcal{I}_0$ to predict the action $a_i$ in each iteration, and then the robot executes the action so that the object would reach state $\mathcal{S}_i$. A new scan is obtained and then segmented with respect to the moveable part to obtain $\mathcal{C}_i^s$. The moveable part $\mathcal{R}_i$ is reconstructed and then masked out for the subsequent interaction analysis on the new interactability map $\mathcal{I}_i$. Simultaneously, $\mathcal{C}_0$ is updated by merging the segmentation results, including movable part removal and interior addition, denoted by $\mathcal{C}_0^i$. After all the moveable parts have been reconstructed, the $\mathcal{C}_0^k$, which constitutes the non-articulated parts, including the scanned interior structures, are finally reconstructed to complete the full geometry acquisition.}
\label{fig_overview}
\end{figure*}

Fig.~\ref{fig_overview} shows an overview and running example of our method, which operates {\em fully automatically\/} by a Fetch robot with built-in depth sensors. At the high level, our reconstruction process consists of two repeated phases:
interaction analysis and interaction-driven reconstruction. Given a target 3D object, the robot first acquires an initial scan, resulting in the initial object state, over which an interaction analysis is carried out by a neural network to detect moveable parts and predict the associated motion attributes. Then, iteratively, for each moveable part detected, the robot executes the 
predicted part manipulation, e.g., opening a drawer as shown in Fig.~\ref{fig_overview}, to reach a new object state for a new point cloud scan and the ensuing mesh reconstruction of the moveable part. In our work, the mesh reconstruction is also performed by a neural network.
Once a moveable part has been reconstructed, it is masked out so that the next interaction analysis would be carried out over the remaining regions of the target object. After all the moveable parts have been manipulated and reconstructed, e.g., the cabinet drawer and then the door as shown in the example in Fig.~\ref{fig_overview}, the robot-object interaction terminates. In the final step, the remaining, non-articulated parts, including all the interior structures that had been exposed by previous part manipulations and subsequently scanned, e.g., the shelving in Fig.~\ref{fig_overview}, are reconstructed.

We conduct experiments on our reconstruction method in both a simulated environment and real-world setup. Qualitative and quantitative evaluations are provided to demonstrate the overall performance of our 3D reconstruction, in terms of quality and completeness of the final meshes and generality across different object categories. Comparisons and ablation studies are also presented to assess the individual components in our solution pipeline.

\section{Related Work}
\label{sec:rw}

3D reconstruction has been one of the most intensively studied problems in both computer graphics and computer vision.
We refer the readers to surveys on classical~\cite{berger2017survey} and learning-based~\cite{NF_survey} reconstruction 
methods for a comprehensive coverage. Among the latter, we only mention neural dual contouring (NDC)
\cite{chen2022ndc}, a state-of-the-art method which we adopt as our mesh reconstruction network due to its versatility, generalizability, and reconstruction quality.
In this section, we mainly cover related works on active 3D reconstruction, neural perception and reconstruction 
of part articulations, as well as simulation environments that support agent-object interactions.

\vspace{-5pt}

\paragraph{Active 3D reconstruction}
Classical 3D acquisition requires continuous scanning around a 3D object, e.g., in Kinect fusion~\cite{newcombe2011kinect}, where the viewpoints and scanning paths are pre-determined.
Yan et al.~\shortcite{yan2006automatic} propose an algorithm for building the kinematic chain of an articulated object from feature trajectories, which first segments the trajectories by local sampling and spectral clustering, and then builds the kinematic chain from a graph constructed from the segmented motion subspaces.
Katz et al.~\shortcite{katz2013interactive} present an interactive perceptual skill for segmenting, tracking, and modeling the kinematic structure of 3D articulated objects.
Representative examples of active 3D scanning, sometimes referred to as autoscanning, include~\cite{wu2014qualitydriven} and \cite{yang2018active}.
Both works develop guided view planners to optimize viewpoint selection and their scanning paths, but neither allows direct manipulation of the target object being acquired.

Despite the vast literature on 3D reconstruction, very few works have been devoted to the capture of object interiors.
Most closely related to our work is {\em proactive scanning\/}~\cite{yan2014proactive}, which executes continuous 3D scanning
while a human user actively interacts with the captured scene, e.g., opening the drawer of a cabinet or the trunk of a car, to access occluded regions.
However, there are a few downsides of this approach: a) it requires substantial {\em human\/} effort and time to obtain a complete reconstruction; 
b) it imposes several technical challenges involving motion perception and tracking, especially due to the presence of 
both human and camera movements, on top of scene modification. For example, the human user is an added occluder and an entity 
that also exhibits motions that must be separated from part articulations by the target object.

In contrast, our reconstruction method is fully automatic, replacing the human user by a robotic agent to carry out the part manipulation and active scanning 
and reconstruction. With our current acquisition setup, both the robot and camera movements can be precisely controlled. On the other hand, the
technical challenges are now shifted to interactivity analysis and part manipulation.

\vspace{-5pt}

\paragraph{Neural perception and reconstruction for articulated shapes}
Yi et al.~\shortcite{yi2018deep} extract the motion type and joint parameters for articulated parts given point cloud shapes with two different articulation states. 
RPM-Net~\cite{yan2019rpmnet} and Shape2Motion~\cite{wang2019shape2motion} are two frameworks which directly infer the motion attributes associated with object parts from a single point cloud frame.
Xu et al.~\shortcite{kawana2022unsupervised} propose an unsupervised framework that learns consistent part parsing for man-made articulated objects with various part poses from a single-frame point cloud.

There has also been recent research on direct inference of interactions. 
Martín-Martín et al.~\shortcite{martin2016integrated} propose an integrated approach for pose tracking, shape reconstruction, and the estimation of kinematic structures, where the three tasks complement each other.
Paolillo et al.~\shortcite{paolillo2017visual} present a method based on virtual visual servoing to estimate the configuration of articulated objects.
Where2Act~\cite{mo2021where2act} and AdaAfford~\cite{wang2022adaafford} are two deep-learning based methods that are trained to predict correct action parameters for object part manipulation. 
Our work adopts Where2Act to construct the interactability map.

In terms of representation learning and neural reconstruction for articulated shapes, most recent works resort to implicit representations \cite{mescheder2019occupancy,chen2019imnet,park2019deepsdf} due to their superior representation capabilities for continuous functions.
Mu et al.~\shortcite{mu2021asdf} introduce A-SDF to represent articulated shapes with a disentangled latent space. Given an unseen shape instance in a random articulation state, it is able to generate novel states for the shape.
Zhang et al.~\shortcite{zhang2021strobenet} present StrobeNet that reconstructs animatable 3D models of articulated objects from one or more unposed RGB images.
In NASAM, Wei et al.~\shortcite{wei2022self} learn a neural shape and appearance model for articulated shapes in a self-supervised manner. Ditto~\cite{jiang2022ditto} is a framework that jointly predicts part-level geometry and joint parameters given two point cloud shapes exhibiting different articulation states.
Hsu et al.~\shortcite{hsu2023Dittoith} further extend the method to the scene level, which predicts the motion parameters and geometries for the objects in the indoor scene.
Similar to Ditto, Nie et al.~\shortcite{nie2023structure} proposed a method to recover part geometries and joint parameters from a sequence of interactions.
Unlike these methods that propose pure network frameworks and apply them on real platforms, the pipeline of our active reconstruction method involves specific robot operations and makes use of robot feedback.

\vspace{-5pt}

\paragraph{Simulation environments}
VirtualHome~\cite{puig2018virtualhome} is a simulator for users to create an activity video dataset with rich ground-truth data. It is driven by programs controlling agents in a synthetic world.
SAPIEN~\cite{xiang2020sapien} is a realistic, physics-rich simulator which hosts a moderately large set of articulated objects from PartNet-Mobility (see below). It enables various robotic vision and interaction tasks that require detailed part-level understanding.
iGibson~\cite{shen2021igibson, li2022igibson} is a simulation environment that aims to develop robotic solutions for interactive tasks in large-scale realistic scenes. It contains several fully interactive indoor scenes populated with both rigid and articulated objects. In our experiments, we utilize iGibson as the virtual setup.

\vspace{-5pt}

\paragraph{3D datasets}
3D shape collections play a key role to enrich the simulation environments. ShapeNet~\cite{chang2015shapenet} is the largest (more than 3M models) and most widely adopted dataset for shape analysis and geometric deep learning. However, there is no joint information between the object parts in ShapeNet, which makes these models hard to use in such tasks as part-level interaction in robotics and VR/AR. PartNet~\cite{mo2019partnet} is a dataset of 3D shapes annotated with instance-level and hierarchical part information.
PartNet-Mobility is an asset of the SAPIEN~\cite{xiang2020sapien} simulator; it contains 14,000 parts from 2,346 object models, many of which do possess interior structures.
Liu et al.~\shortcite{liu2022akb48} introduce AKB-48, which is a large-scale articulated object knowledge base that consists of 2,037 real-world 3D articulated object models across 48 categories.

\section{Overview}
\label{sec:overview}

Given a target 3D object $\mathcal{S}_0$ to be fully reconstructed, from the outside to the inside, our Fetch robot first moves in front of the object to obtain an initial point cloud scan $\mathcal{C}_0$ using its built-in depth camera. Note that we do not perform any reconstruction in this initial step.
The key first task is to analyze the interaction for $\mathcal{C}_0$ and obtain the {\em interactability map\/} $\mathcal{I}_0$. Specifically, the map is defined by a point-wise probability indicating the likelihood that each point belongs to a part that can be successfully manipulated.
More details about the interaction analysis are given in Section~\ref{sec:interaction}.

To enable a part manipulation, we need to pass a specific action to the robot. We denote an action as ${a} = (a^{pos}, a^{dir}) \in \mathbb{R}^6$, where $a^{pos}$ represents the action position and $a^{dir}$ the action direction. Given an action $a$, the robot moves its gripper to $a^{pos}$ and manipulates the part along the direction $a^{dir}$.
Once the part is successfully opened, we reach a new state $\mathcal{S}_1$ for the target object. In addition, the mobility information of the part, including motion axis and motion range, can be obtained by fitting the parameters from the trajectory of the manipulation process. Note that this step is critical as it corrects the predicted action by the actual interaction.

Next, the robot moves back to the front of the current object configuration to obtain a new point cloud $\mathcal{C}_1$, which is further segmented by a network to obtain $\mathcal{C}_1^s$. According to the segmentation label, we complete and reconstruct the current moving part $\mathcal{R}_1$ using a completion and reconstruction module that is trained to perform well amid noise and sparse scans.
The update of $\mathcal{C}_0$ is obtained by removing the moveable part and adding the new observed parts based on segmentation masks, denoted as $\mathcal{C}_0^1$.

Since the target 3D object may have more than one movable part, we update the interactability map by masking out the segmentated part, denoted as $\mathcal{I}_1$. Then, we repeat the action selection, manipulation, scanning, and segmentation process to obtain the point cloud $\mathcal{C}_i^s$, reconstruction of the current part $\mathcal{R}_i$, while the non-articulated point cloud $\mathcal{C}_0^{i}$, $i \ge 2$, can be updated accordingly.
Note that each time before the next action, the robot will first move the last manipulated part back to restore its original configuration.

To determine when the interaction analysis shall terminate, we examine the updated interactability map $\mathcal{I}_k$ to obtain the number of points deemed to be sufficiently moveable. If this number falls below a preset threshold, then the part manipulation would stop, reaching the final completion and reconstruction step, which takes the remaining part $\mathcal{C}_0^k$ as input.
In the end, we merge the reconstructed meshes and the part mobility information to obtain our final active reconstruction $\mathcal{R}$. 
More details about the interaction-driven reconstruction are presented in Section \ref{sec:reconstruction}.

\section{Interaction analysis}
\label{sec:interaction}

\begin{figure}[!t]
\centering
\includegraphics[width=0.85\linewidth]{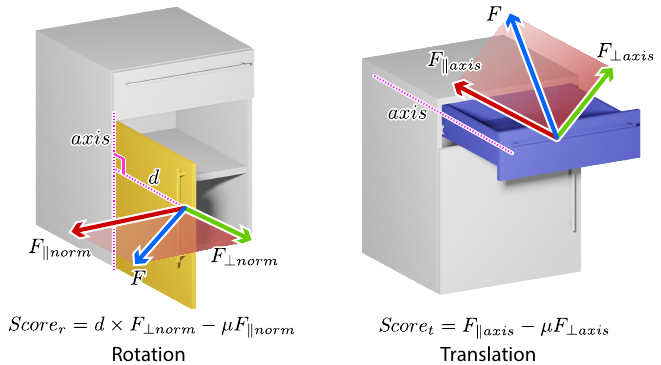}
\caption{The illustration of the rules to generate training data for interaction perception network, where the force $F$ is decomposed in two different ways for each type of motions to compute the interactability scores.}
\label{fig_interaction_rule}
\end{figure}

Given the initial point cloud $\mathcal{C}_0$ of the input shape $\mathcal{S}_0$, the goal of our interaction analysis is to predict the initial interactability map $\mathcal{I}_0$. We first use a network module to give an initial prediction of the interaction, then it is gradually updated at each of the following steps until the stop condition is reached.

\begin{figure}[!t]
\centering
\includegraphics[width=\linewidth]{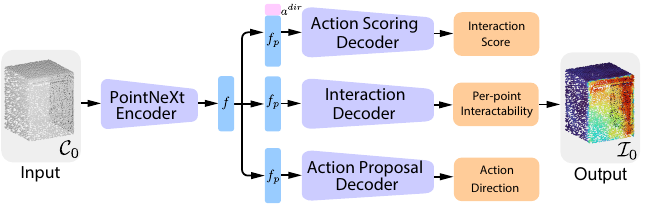}
\caption{The architecture of the interaction perception network. Given the initial scan $\mathcal{C}_0$, the feature $f$ is first extracted, and then the interactability score for each point is predicted, denoted as $\mathcal{I}_0$. The action direction can be inferred from the action proposal decoder, and the action scoring decoder is trained to provide guidance for the other two decoders.}
\label{fig_net_interaction}
\end{figure}

\paragraph{Rule-based training data generation}
We first collect the ground truth (GT) data of how likely an interaction $a = (a^{pos}, a^{dir})$ is able to move the part. 
Where2Act~\cite{mo2021where2act} uses simulation in virtual environment to collect the training data. 
Specifically, they first sample a large number of interactions in 3D space, and then test each of them in SAPIEN~\cite{xiang2020sapien}: if the state of the interacting part after interaction meets their pre-defined success conditions, this interaction will be labeled as true.
However, since all the sampled interactions need to be tested via simulation, it is time-consuming and the binary GT label could not provide detailed supervision for the interactability prediction.

Our key observation is that the interactability of articulated objects is highly related to its part mobility.
For example, the door is more easily opened from the side away from the hinge.
Since each model in our dataset is associated with its part mobility information, it allows us to define several physics-based rules that can be used to calculate the interactability score given an action $a = (a^{pos}, a^{dir})$ on a point cloud shape to serve as GT for training. 

Fig.~\ref{fig_interaction_rule} shows the interactability score computation of two examples with different types of motions.
For the part with rotational motion, given a force $F$ on $a^{pos}$ with direction $a^{dir}$, we decompose the force $F$ along the normal direction at the current position. To obtain the torque, we calculate the smallest distance $d$ from the action position to the axis, and the action score is defined as: 
\begin{equation}
    {Score_{r}} = d\times F_{\perp {norm}} - \mu F_{\parallel {norm}},
\end{equation}
where $\mu$ is the friction coefficient.

For the part with translational motion, we decompose $F$ along with two directions: the direction parallel to the axis $F_{\parallel {axis}}$, which drive the part to move; and the direction perpendicular to the axis $F_{\perp  {axis}}$, which would produce resistance during the movement. We define the score for translation as:
\begin{equation}
    {Score_{t}} = F_{\parallel {axis}} - \mu F_{\perp {axis}},
\end{equation}
where $\mu$ is the friction coefficient.

Note that the score distribution for translation and rotation may be different, so we normalize the score for each movable part to generate the GT training data. 
Specifically, given a point cloud, we randomly generate 10 directions in 3D space for each point with the same segmented label and compute 10 scores. 
Then all the scores of the part for that label are collected and normalized to the $[0, 1]$. 

\paragraph{Network structure.}
The interaction perception network used to predict the initial interactability map $\mathcal{I}_0$ is adapted from the work of Where2Act~\cite{mo2021where2act}, with the network architecture shown in Fig.~\ref{fig_net_interaction}.
The network consists of one encoder and three different decoders, where the action scoring decoder is trained first, then it is used to guide the training for the other two decoders.
The network first extracts the point-wsie feature $f$ of the input $\mathcal{C}_0$, and then for each point $p$, the corresponding feature $f_p$ together with a direction $a^{dir}$ will be passed to the action scoring decoder to infer the success probability.
To use the success probability predicted by the action scoring decoder to guide the training of the interaction decoder for per-point interactability, 
for each point, we sample 100 directions $a^{dir}$  and pass them to the action scoring decoder to get 100 scores, and then we use the average score as the GT interactability score.
Moreover, we train the third action proposal decoder to predict the optimal interaction direction for each point with the highest success score to provide full guidance for interaction manipulation of the robot.
More details about the network parameters can be found in the supplementary material.

\paragraph{Loss function.}
The loss function for the action proposal decoder $\mathcal{L}_i^p$ is the cosine distance between the predicted and GT action directions.
For the action scoring decoder, differently from Where2Act which treats the problem as that of a binary classification, 
we calculate the score for each action based on the rules, and thus, the loss function $\mathcal{L}_i^s$ is defined as the mean square error between the predicted and GT scores.
For the interaction decoder, the loss function $\mathcal{L}_i^a$ is defined as the mean square error between the predicted and the average score for 100 action samples obtained from the action proposal and action scoring decoders.
The final loss for the interaction prediction network is then:
\begin{equation}
    \mathcal{L}_i = \omega_d \mathcal{L}_i^p + \omega_s \mathcal{L}_i^s + \omega_a \mathcal{L}_i^a,
\end{equation}
where we set the weights $(\omega_d, \omega_s, \omega_a)$ to be $(1, 1, 50)$.

\section{Interaction-driven reconstruction}\label{sec:reconstruction}

\subsection{Interaction execution}
Our active reconstruction method works automatically by using a Fetch robot to execute a set of operations: mobile scanning, action selection, part manipulation, and termination.
The details of each operation are described below.

\paragraph{Mobile scanning}
Given a shape in front of the robot, we first estimate its bounding box based on the captured depth image and then generate two new viewpoints to capture two new depth images from the two front sides.
The point cloud data is then obtained by fusing the depth maps together with the corresponding camera parameters. Note that the normal direction for each point is also calculated according to the camera extrinsic.

\paragraph{Action selection}
Given an interactability map $\mathcal{I}$, we take the action candidate with the highest probability to guide the part manipulation. Then, the robot attempts to manipulate the object part following the action. 
If the action fails to move the part, another action candidate will be selected from $\mathcal{I}$. Specifically, to avoid the candidates falling in a small region of the part, we first filter all the points around the failed action point with radius $r$, and then sample another action candidate with the highest probability in the remaining region. 
We set $r = 0.05 m$ in our experiments.

\paragraph{Part manipulation}
For the execution of part manipulation, we replace the default clamp-based gripper on the Fetch with the suction-based gripper, so that the robot can perform any interaction without knowing the specific action type, such as pushing, pulling, etc.
Given an action $(a^{pos}, a^{dir})$, the gripper first approaches the position $a^{pos}$. 
To make sure the suction state is stable, the gripper would move forward an additional $0.01 m$ along the inverse direction of $a^{dir}$ before starting suction in our experiments.
During the manipulation process, the robot captures the point cloud at every time step $t=3s$ and compares it to the initial state to see if the part has been moved or still moving. 
As the robot location and camera are fixed during the manipulation, we calculate the Chamfer distance between two point clouds, and if the distance is smaller than $1.5e^{-3}$, the release signal will be sent to the gripper to stop the manipulation action.

\paragraph{Termination.}
The stopping condition for robot operations is:
\begin{equation}
    {Count}(\forall a \in \mathcal{I} > T_a) < T_c
\end{equation}
where $T_a = 0.8$ is an action score threshold and $T_c = 30$ the action count threshold. 
This condition is checked each time when an updated interactability map is received.

\begin{figure}[!t]
\centering
\includegraphics[width=0.9\linewidth]{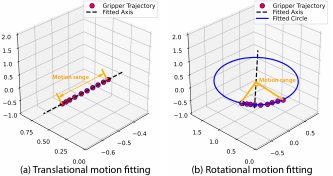}
\caption{The examples of the motion fitting. Given a set of points sampled from the gripper trajectory, we fit the motion axis and motion range using least squares optimization.}
\label{fig_axis_fit}
\end{figure}

\begin{figure*}[!t]
\centering
\includegraphics[width=\linewidth]{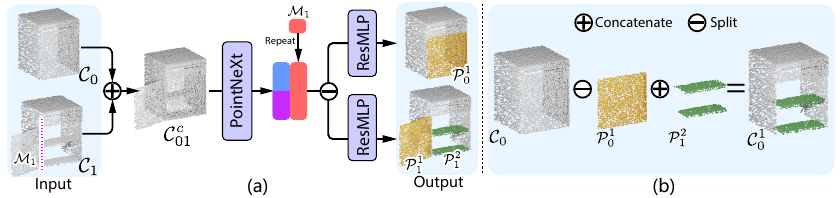}
\caption{The architecture of the part segmentation network (a) and the update process for the non-articulated parts (b). Given the initial point cloud $\mathcal{C}_0$ and current observation $\mathcal{C}_1$ with the part motion $\mathcal{M}_1$, the network predicts the segmentation for the movable parts $\mathcal{P}_0^1$ and $\mathcal{P}_1^1$ in initial and current states, and interior of the non-articulated part $\mathcal{P}_1^2$. The updated point cloud $\mathcal{C}_0^1$ is obtained by masking out $\mathcal{P}_0^1$ from $\mathcal{C}_0$ and further concatenate with $\mathcal{P}_1^2$.}
\label{fig_net_segmentation}
\end{figure*}

\subsection{Active reconstruction}
Our active reconstruction utilizes not only scans of the object in different states to get the geometrical surface but also the moving trajectory of the gripper to estimate the motion parameters for each part. Moreover, the part motion parameters are used to help with the movable part segmentation, and thus we first give details about the motion acquisition and then explain the part segmentation and reconstruction modules. 
Note that as the movable parts are manipulated one by one during the whole active reconstruction process, the point cloud of each part is segmented and reconstructed separately in a row, which will eventually be combined together to form the final complete shape reconstruction.

\paragraph{Motion acquisition}
Although the action predicted by the interaction perception network may not be perfectly accurate, the movement of the object part during manipulation is always unique.
Therefore, we record the movement trajectory of the robot gripper during the interaction process, and fit the motion parameters, including motion axis, motion range and motion type, as illustrated in Fig.~\ref{fig_axis_fit}.

We start recording the position of the gripper once the start suction signal is sent to the robot.  
We then uniformly sample 20 positions following the moving trajectory, and use least squares optimization to fit either a circle or a line. 
To judge whether the motion type is rotation or translation, we first fit a circle given the position samples, and if the radius of the circle is larger than a threshold $T_r$, we set the motion type to translation. We set $T_r$ to 1 in our experiments, which is the size of normalization bounding box.

For rotation, the axis of motion is the line perpendicular to the fitted circle at the center.  
We draw two lines from the start and end positions to the circle center, and use the angle between the lines to define the rotation range. 
For translation, the axis is the fitted line itself, and the range is inferred by computing the distance between the start and end positions. 
For each motion type, we represent axis as $\mathcal{M} = (x, y, z, u, v, w)$, where $(x, y, z)$ is the axis position and $(u, v, w)$ is the axis direction. 
The rotation and translation ranges are represented as an angle value and length value, respectively.
In this way, we obtain the exact motion information for each moving part by fully utilize the manipulation feedback.

\paragraph{Part segmentation}
Given the initial point cloud $\mathcal{C}_0$, the one after robot manipulation $\mathcal{C}_1$, together with the part motion parameters vector $\mathcal{M}_{1}$, we design a part segmentation network to extract the corresponding moving part as well as the newly detected parts, as shown in Fig.~\ref{fig_net_segmentation}.
Unlike most existing networks that take individual point cloud shapes as input and segment all the movable parts, our goal is to recognize the variable region of a pair of shapes before and after the interaction.
Thus, we directly concatenate the two point clouds so that the points density of the fixed region is higher than the region that is changed, which could be easier for the network to capture and thus lead to better results.

In more detail, those two point clouds with shape 4,096$\times$3 are first concatenated into a new point cloud $\mathcal{C}_{01}^c$ with shape 8,192$\times$4,
where the fourth dimension stores the label for each point indicating whether the point comes from $\mathcal{C}_0$ or $\mathcal{C}_1$.
Then, we pass $\mathcal{C}_{01}^c$ to the PointNeXt encoder~\cite{qian2022pointnext} to extract the point-wise feature $f$ with shape 8,192$\times$64.
The feature is then concatenated with the motion parameters $\mathcal{M}_{1}$ with shape 1$\times$7. Note that the motion vector is repeated 8,192 times to concatenate with $f$. Then we split the concatenated feature and pass each of them to two Residual MLP decoders to predict the masks corresponding to the segmentation on each of the input $\mathcal{C}_0$ and $\mathcal{C}_1$.
Parts $\mathcal{P}_0^1$, $\mathcal{P}_1^1$ shown in yellow represent the moving part in either the initial or the manipulated scans, while $\mathcal{P}_1^2$ shown in green represents the interiors of the non-articulated part observed after robot manipulation.
The loss function for part segmentation network is defined as the cross entropy loss between the predicted and GT segmentation labels.

Note that our segmentation module not only segments the movable parts but also discovers the inner structure of the non-articulated part, which can be used to update the initial point cloud $\mathcal{C}_0$.
Specifically, $\mathcal{C}_{0}$ first masks out the moving part $\mathcal{P}_0^1$ and then is concatenated with the interior $\mathcal{P}_1^2$. We denote the updated shape as $\mathcal{C}_0^i$, with an example for iteration $i=1$ shown in Fig.~\ref{fig_net_segmentation}(b).

\begin{figure}[!t]
\centering
\includegraphics[width=\linewidth]{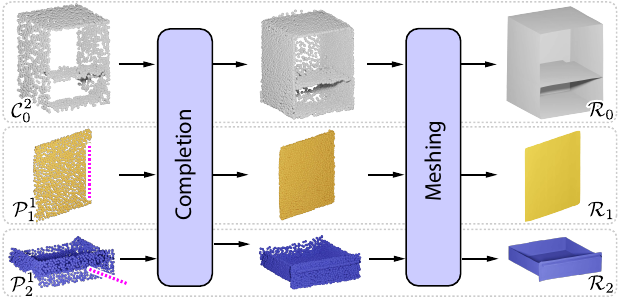}
\caption{The completion and reconstruction module of our method, where the completion network is adapted from SeedFormer~\cite{zhou2022seedformer} and pre-trained NDC~\cite{chen2022ndc} is directly used for meshing.}
\label{fig_net_reconstruction}
\end{figure}

\begin{figure*}[!t]
  \includegraphics[width=\linewidth]{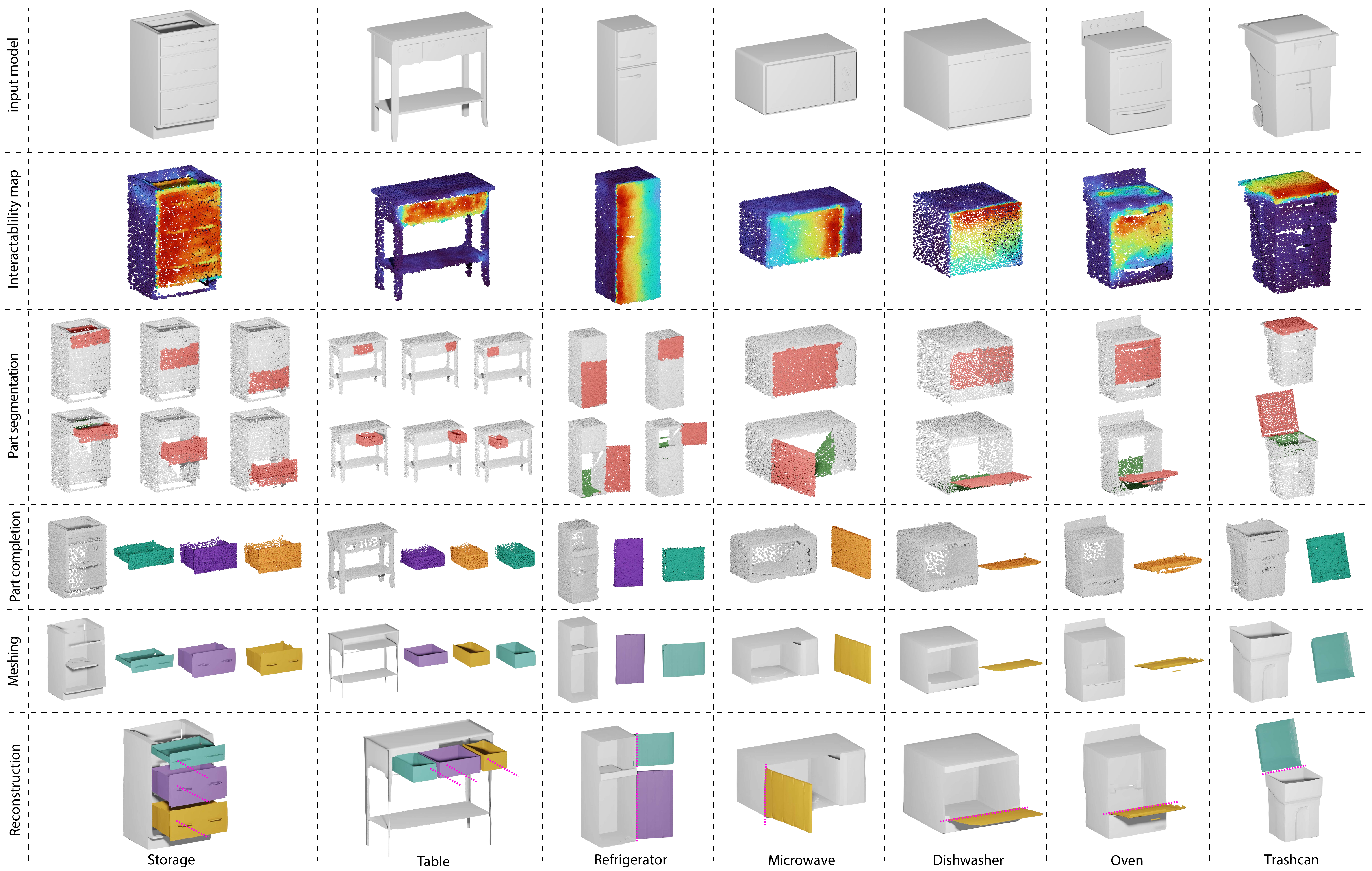}
  \caption{The reconstruction results of our method on synthetic data. The input model is shown in the first row, and the interactability maps predicted by the interaction perception network are shown in the second row, where the red color indicates higher interactability.
  In the next row, we show the part segmentation masks, where the results corresponding to the initial object states and the manipulated object states are shown in two rows inside the cell, respectively. 
  The mask of the moving part is drawn in red color, while the newly-observed interior of the non-articulated part is shown in green color. 
  The completion result and the corresponding reconstruction mesh for each part are shown in the fourth and fifth row. 
  The last row demonstrates the final reconstruction of our method, with the motion axis drawn in magenta for each of the movable parts.}
  \label{fig_gallery}
\end{figure*}

\paragraph{Part reconstruction}
Fig.~\ref{fig_net_reconstruction} shows an example of our reconstruction pipeline that takes segmented parts as input and generates the complete mesh.
Due to self-occlusion, most of the part scans are incomplete even if they are fully opened, so we first complete each of the parts and then reconstruct them.
Specifically, each part together with its motion information is fed into a completion network adapted from SeedFormer~\cite{zhou2022seedformer} to predict the missing region.
Note that for the non-articulated part, its associated motion parameters are all set to zero.
The loss functions for completion are the same as the original SeedFormer.
We then feed the completed point cloud to NDC~\cite{chen2022ndc}, which is the state-of-the-art meshing algorithm that generates high-quality mesh from a variety of inputs including point clouds.
Note that as the original NDC is trained on a large dataset with nice generalizability, we directly use the pre-trained NDC model to get the reconstruction results.

\section{Results and Evaluation} \label{sec:results}

\subsection{Experimental Setup}
\paragraph{Simulation environment}
We use iGibson~\cite{li2022igibson} as the simulation environment to develop the whole interaction 3D reconstruction method. We initialize the scene with an indoor room without furniture. Then, we load each of the models to a specific position in the middle of the room and perform our method.

\paragraph{Real world setup}
We use a Fetch Mobile robot which has a 7-DoF arm to conduct a series of operations in real-world scenarios. We add a suction module to the original gripper of Fetch. The suction is controlled by a vacuum pump and an air release valve, both of which are electrical. This allows the robot to control the state of the suction gripper by sending signals to the pump.
For more details about the robot settings, please refer to the supplementary.

\paragraph{Dataset}
We select seven categories from the PartNet-Mobility dataset~\cite{xiang2020sapien} to train and test our method, with a total of 434 shapes and 827 parts. The train/test ratio is set to 8:2, and all the quantitative evaluation and comparisons are based on the same test set.
Each of the shapes $S$ in the dataset is stored in URDF (Unified Robot Description Format) format, which specifies the attributes (kinematic tree, names, ranges, etc.) of the joints and links of the shape.
Given a URDF shape model, we extract each of the part models along with the motion axis, motion type, and motion range. Then, we normalize each shape using a unit box located at $[0.5, 0.5, 0.5]$, and each part of the shape is transformed based on its motion axis and motion range information. Afterward, we obtain the point cloud data by scanning the models from three viewpoints, which are located in front of the object to simulate the robot view.
More details about the training data generation process are provided in the supplementary.

\subsection{Qualitative results} \label{sec:qualitative}

\paragraph{On synthetic data.}
Fig.~\ref{fig_gallery} shows one example reconstruction result for each category in our dataset.
We can see that our active reconstruction obtains high-quality results in each of the steps. For interaction perception, our network is able to distinguish between the single door and double side doors with similar geometry of storage and refrigerator. For microwave, dishwasher, and trashcan, higher probability accurately distributed in the correct side of the shape, corresponding to the rotations of side-opening, down-opening, and up-opening, respectively. 
For part segmentation, our network is able to predict the correct mask for a pair of point cloud shapes in initial and manipulated states. 
Moreover, the interior of the non-articulated parts is also accurately segmented.
From the fourth row, we can see that our method can generate complete and reasonable geometries for the partial point cloud, and the following reconstruction meshes for each part are also clean and well-structured for movable parts and non-articulated parts. For example, we obtain the complete geometry for the drawers of the table, and the body of the storage with the clapboards is also fully recovered.
The last row demonstrates the final reconstruction of our method, with the correct motion axis for each of the movable parts.

\begin{figure}[!t]
  \includegraphics[width=0.95\linewidth]{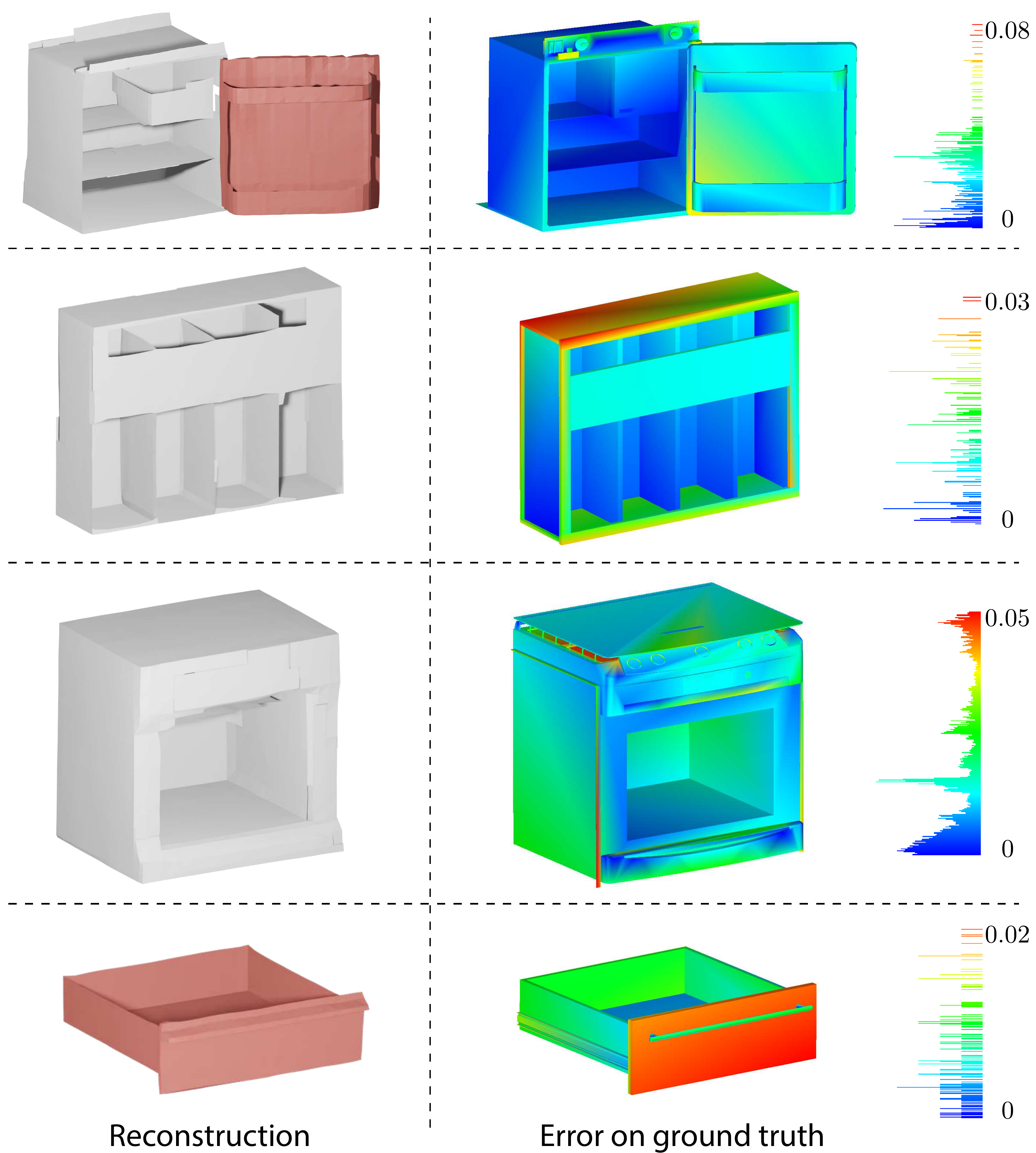}
  \caption{Reconstruction quality for our results indicated by the error map shown on the ground truth models. The error is calculated using the Hausdorff distance, with larger distances colored in red.}
  \label{fig_quality}
\end{figure}

To further illustrated the reconstruction quality of the interior parts, we show the error map for several reconstructions with respect to their corresponding GT results in Fig.~\ref{fig_quality}.
Specifically, we calculate the Hausdorff distance between the vertices sampled on the two meshes, with higher error colored in red. The range of error values for each histogram is labled on the right.
The first row shows a refrigerator reconstructed by our method, and we can see both the door and the interior region of the refrigerator are reconstructed accurately.
The second and third rows show another two examples of the reconstruction of the non-articulated part. Note how they are nicely captured and reconstructed with small errors.
For the oven shown in the Third row, the region with higher errors is mainly distributed on the heat dissipation area of the oven, with is occluded by the cover.
To better illustrate the reconstruction of specific moving parts, we show a reconstructed drawer of a cabinet in the last row, and we can see that the thin structure like the handle is preserved on the reconstruction, and the error region on the front surface are higher since the location of the reconstructed surface are slightly shifted comparing to the GT.

\begin{figure*}[!t]
  \includegraphics[width=\linewidth]{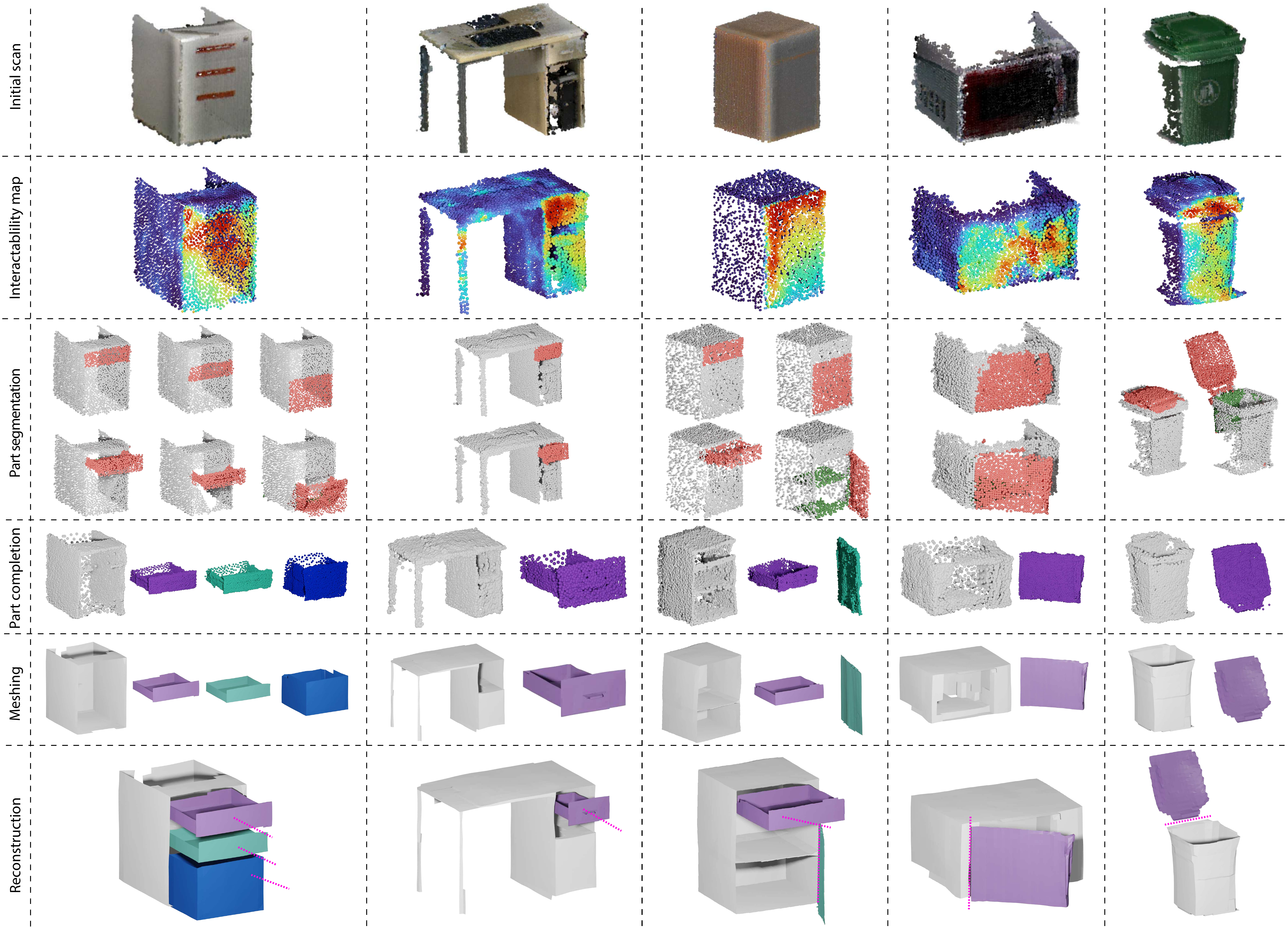}
  \caption{The reconstruction results of our method on real scanned objects. We show the initial scan and predicted interactability in the first two rows, the results of part segmentation, completion, meshing, and the final reconstruction are shown in the following rows.}
  \label{fig_gallery_real}
\end{figure*}

\paragraph{On real data.}
Fig.~\ref{fig_gallery_real} shows some example reconstruction results on real scanned objects.
The first column shows a storage with three drawers, whose whole reconstruction pipeline is also shown in the supplementary video.
The objects in the following four columns are coming from the same indoor scene, as shown in Fig.~\ref{fig:teaser}. The robot first scan in front of the scene, then we segment each of the objects, so that the robot knows the location of each object, and then performs the reconstruction process respectively.
We can see that the point clouds scanned in the real world are incomplete and noisy, while our method works robustly and can successfully predict interactability, manipulate parts with the robot arm, and reconstruct the geometry with accurate motion parameters.

\subsection{Quantitative evaluation} \label{sec:quantitative}
In this section, we first introduce the quantitative metrics and then compare our active reconstruction method to state-of-the-art works.
We also conduct several ablation studies to evaluate each module in our method and justify the design choices.

\paragraph{Evaluation metrics.}
To measure the accuracy of the interaction perception network, we count the number of action attempts for reconstructing each of the shapes.
Specifically, given a test shape, we run our method and count the number of action candidates $M = \sum_{n = 1}^{k} {Count}(a_{candidate}^n)$ selected for robot manipulation during the whole process, where $k$ is the number of movable parts. In the best situation, $M$ would be equal to the number of moving parts, which means that all the parts are successfully manipulated by the first attempt of the robot. We denote the accuracy of the interaction perception network as $A_{action} = \frac{k}{M}$.

For the part segmentation network, we measure the accuracy for each shape as $A_{seg} = \frac{p_{correct}}{p_{total}}$, where $p_{correct}$ is the number of points with predicted labels matching the GT labels, and $p_{total}$ is the number of points for each shape.
We also use the mean IoU to measure the prediction accuracy, denoted as $mIoU$.
Since we segment the point clouds in both the initial state and the manipulated state, the accuracy is calculated for each state respectively, denoted as $A_{seg}^{start}$, $A_{seg}^{end}$, ${mIoU}^{start}$, and ${mIoU}^{end}$.

For the completion network, we measure the Chamfer Distance (CD) $E_{comp}^{cd}$ and Earth Mover's Distance (EMD) $E_{comp}^{emd}$ between the generated and GT point clouds.

For the reconstructed mesh, we first unifomly sample $2048$ points on the surface of predicted and ground truth mesh, then calculate CD and EMD between the sampled points to measure the reconstruction error, denoted as $E_{recon}^{cd}$ and $E_{recon}^{emd}$.

For part motion acquisition, including motion axis and range, we measure the error of axis direction $E_{mo}^{dir}$ by calculating the cosine between the predicted and GT direction; the error of axis position $E_{mo}^{pos}$ is the L2 distance between the predicted and GT position. Note that for the translation type of motion, the position of the axis is less important, thus we only calculate $E_{mo}^{dir}$ for translation.

\begin{table*}[!t]
	\caption{Quantitative results of our method. Note that $E_{comp}^{cd}$, $E_{recon}^{cd}$ are multiplied by $10^3$, while $E_{comp}^{emd}$, $E_{recon}^{emd}$, $E_{mo}^{dir}$ and $E_{mo}^{pos}$ are multiplied by $10^2$.}
	\label{tab_quantitative}
	\begin{tabular}{c||c|c|c|c|c|c|c||c} \hline
	Category & Storage & Table & Refrigerator & Microwave & Dishwasher & Oven & Trashcan & Average \\ \hline
	$A_{action} $ & 0.21 & 0.25 & 0.25 & 0.68 & 0.58 & 0.38 & 0.58 & 0.41  \\ \hline
	${mIoU}^{start}$  & 0.964 & 0.924 & 0.950 & 0.934 & 0.970 & 0.952 & 0.937 & 0.957 \\ \hline
	${mIoU}^{end}$    & 0.970 & 0.928 & 0.952 & 0.964 & 0.970 & 0.941 & 0.923 & 0.963 \\ \hline
	$A_{seg}^{start}$ & 0.991 & 0.994 & 0.982 & 0.976 & 0.987 & 0.981 & 0.976 & 0.989 \\ \hline
	$A_{seg}^{end}$   & 0.991 & 0.997 & 0.983 & 0.984 & 0.988 & 0.981 & 0.974 & 0.990 \\ \hline
	$E_{comp}^{cd}$   & 0.273 & 0.313 & 0.213 & 0.217 & 0.153 & 0.351 & 0.257 & 0.254 \\ \hline
	$E_{comp}^{emd}$  & 1.624 & 2.626 & 1.158 & 1.343 & 1.095 & 1.726 & 1.507 & 1.583 \\ \hline
	$E_{recon}^{cd}$  & 0.633 & 0.192 & 0.229 & 0.379 & 0.294 & 0.872 & 0.535 & 0.448 \\ \hline
	$E_{recon}^{emd}$ & 1.074 & 0.457 & 0.699 & 1.096 & 1.752 & 0.881 & 1.005 & 0.999 \\ \hline
	$E_{mo}^{dir}$  & 2.07 & 2.22 & 2.51 & 1.97 & 1.31 & 2.21 & 1.73 & 2.00 \\ \hline 
	$E_{mo}^{pos}$  & 1.12 & 0.86 & 0.87 & 1.08 & 0.98 & 0.54 & 1.32 & 1.13 \\ \hline
\end{tabular}
\end{table*}

\paragraph{Quantitative results.}
The quantitative results of our method are shown in Table~\ref{tab_quantitative}. 
Note that our method works well for all seven categories tested in our experiments.
For interaction perception, we can see that the accuracy of the categories with more parts such as storage and table is lower than that of the object categories that have fewer parts, due to the accumulated errors in robot navigation and localization. 
For segmentation, the accuracy for the end state is slightly higher since the moved region is more distinguishable than the start state. 
The errors for both part completion and mesh reconstruction are low, even for the categories with complex geometries and variations such as table.
Moreover, we can see that the motion parameters predicted by our axis fitting strategy have robust performance with small variance across all categories.

\begin{table}[!t]
	\caption{Quantitative comparisons to A-SDF and Ditto. Note that for A-SDF, we only measure the reconstruction error due to the lack of segmentation and motion output. $E_{recon}^{cd}$, $E_{recon}^{emd}$ are multiplied by $10^3$, $10^2$, respectively.}
	\label{tab_compare_ditto}
	\begin{minipage}{\columnwidth}
		\begin{center}
		\begin{tabular}{c||c|c|c|c|c|c} \hline
			           & $mIoU^{end}$ & $A_{seg}^{end} $ & $E_{recon}^{cd}$ & $E_{recon}^{emd}$ & $E_{mo}^{dir}$ & $E_{mo}^{pos}$ \\ \hline
			 A-SDF     & -       & -      & 4.377 & 10.436 & - & - \\ \hline
			 Ditto     & 0.935   & 0.978  & 0.861 & 3.308  & 9.68 & 5.21  \\ \hline
			 \textbf{Ours} & 0.963  & 0.990  & 0.448 & 0.999 & 2.00 & 1.13 \\ \hline
		\end{tabular}
		\end{center}
	\end{minipage}
\end{table}

\begin{figure}[!t]
  \includegraphics[width=\linewidth]{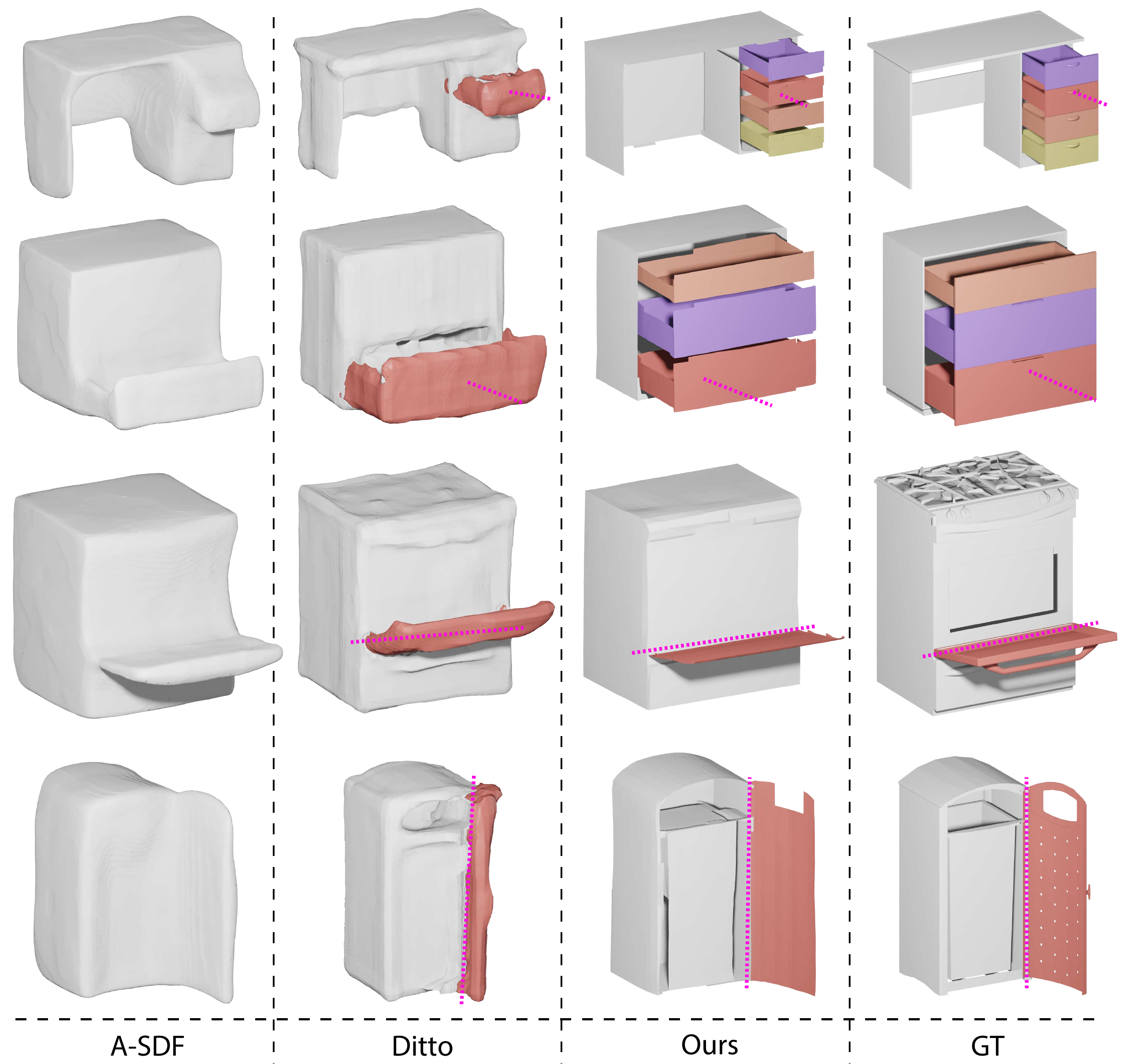}
  \caption{Qualitative comparisons of our method to A-SDF and Ditto. Note that A-SDF reconstructs the whole shape without part segmentation, while Ditto reconstructs shape with only one moving part recovered.}
  \label{fig_compare_ditto}
\end{figure}

\paragraph{Comparison with the state-of-the-art methods}
We mainly compare our method to the two most related state-of-the-art methods Ditto~\cite{jiang2022ditto} and A-SDF~\cite{mu2021asdf}, which predicts joint parameters and geometries for articulated shapes from a pair of shapes in different articulation states. The key difference between the two methods is that we fully utilize the feedback during interaction such as the motion axis, which can enhance the ensuing reconstruction. Also, our method is able to reconstruct shapes with multiple parts automatically. Note that we make the comparison to Ditto on their dataset.

Table~\ref{tab_compare_ditto} shows the quantitative comparisons and some visual comparisons are presented in Fig.~\ref{fig_compare_ditto}.
For visual results, compared to A-SDF and Ditto, our reconstructions can capture more detailed thin structures.
Moreover, the parts reconstructed by our method are more distinguishable, instead of blending with the other parts.  
Our method is also able to reconstruct objects with multiple parts automatically. 
Note how the drawers are successfully recovered in the first two rows, and the inside structure of the trashcan in the last row is also successfully reconstructed.

For the quantitative comparisons, the reconstruction error of our method is lower than the baselines. For A-SDF, it generates the whole shape in different articulate states given a mesh input, and it cannot predict part segmentation and motion information. Compared to Ditto, the part segmentation accuracy of our method is slightly higher, while the motion predicted by our method through trajectory fitting is more accurate than the network-predicted results from Ditto. Note that Ditto predict segmentation on the generated part, so we only show the metrics of $A_{seg}^{end}$ and ${mIoU}^{end}$.

\subsection{Ablation studies} \label{sec:ablation}
To justify the network structures and loss functions designed in our method, we perform several ablation studies.

\paragraph{Design choices of interaction perception network}
We compare our adopted interaction perception network to the original method of Where2Act~\cite{mo2021where2act} in Table~\ref{tab_compare_w2a}. 
$T_{gen}$ denotes the time for generating one action sample, while $T_{test}$ denotes the time for predicting an action given a shape during testing.
Specifically, in Where2Act, the action position and direction are sampled randomly and simulated in the environment to obtain whether the action is successful or not. While in our rule-based data sampling strategy, given a point cloud shape with its motion information, we can directly calculate the scores given arbitrary action, so that the required time is 10,000 times less than Where2act. Also, the time for testing one shape with our method is also faster.
For action prediction accuracy, our method is 2 times higher than Where2Act.

\begin{table}[!t]
	\caption{Quantitative comparisons of interaction perception network.}
	\label{tab_compare_w2a}
	\begin{minipage}{\columnwidth}
		\begin{center}
		\begin{tabular}{c||c|c|c} \hline
			               & $A_{action}$  & $T_{gen}$ (second) & $T_{test}$ (second) \\ \hline
			 Where2Act     & 0.16 & 18.044 & 1.203 \\ \hline
			 \textbf{Ours} & 0.41  & $ 0.0012 $  & 0.251 \\ \hline
		\end{tabular}
		\end{center}
	\end{minipage}
\end{table}

\begin{table}[!t]
	\caption{Quantitative comparisons of part segmentation network.}
	\label{tab_compare_seg}
	\begin{minipage}{\columnwidth}
		\begin{center}
		\begin{tabular}{c||c|c|c|c} \hline
			                  & $mIoU^{start}$ & $mIoU^{end}$ & $A_{seg}^{start}$ & $A_{seg}^{end}$  \\ \hline
			 Seg-baseline    & 0.928          & 0.934        & 0.963             & 0.971   \\ \hline
			 Ours w/o motion  & 0.951          & 0.962        & 0.982        & 0.988  \\ \hline
			 \textbf{Ours}  & 0.957          & 0.963        & 0.989          & 0.990   \\ \hline
		\end{tabular}
		\end{center}
	\end{minipage}
\end{table}

\begin{figure}[!t]
  \includegraphics[width=\linewidth]{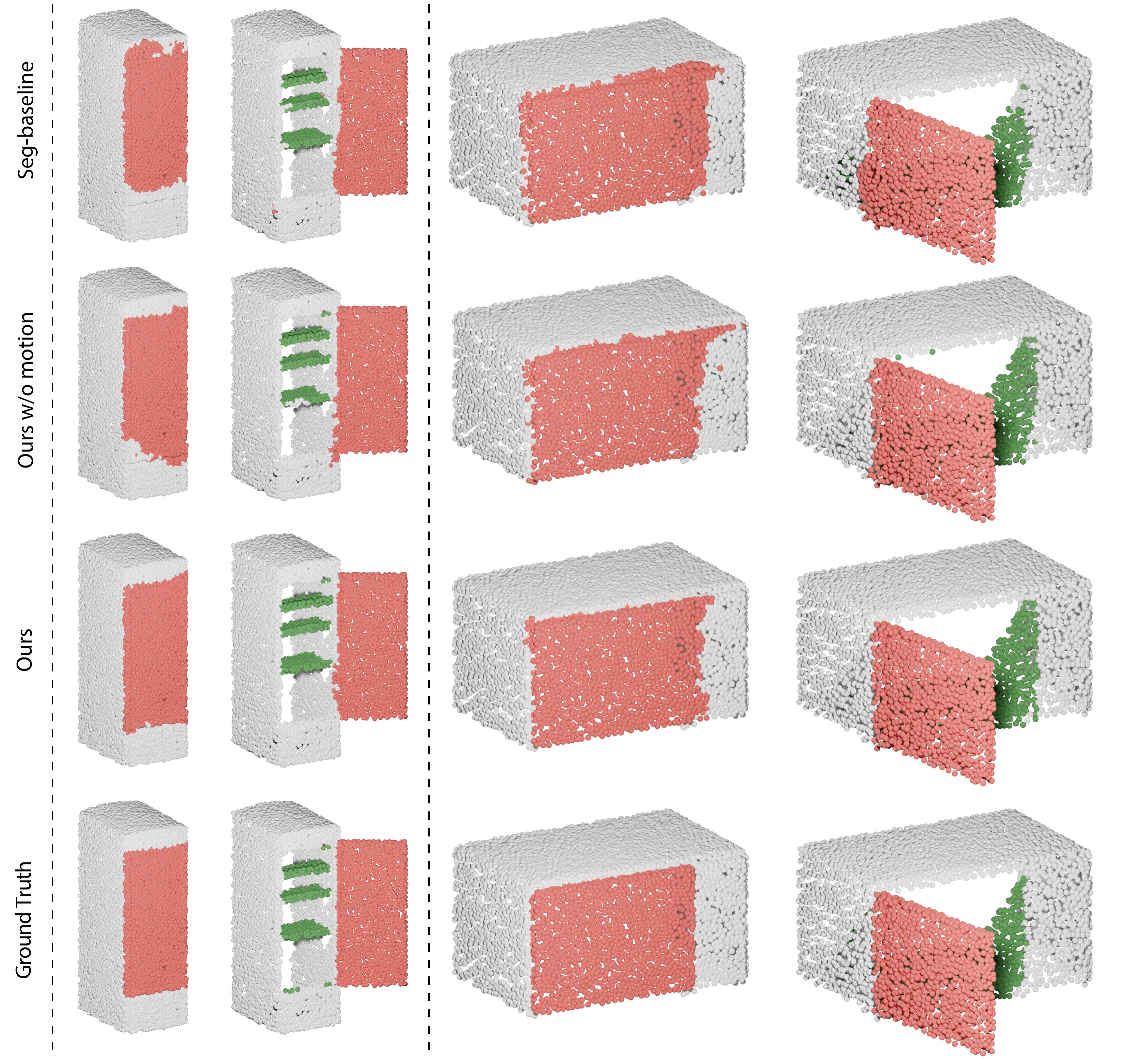}
  \caption{Qualitative comparisons of our segmentation network to the baselines. The mask of the moving part is drawn in red color, while the newly-observed interior of the non-articulated part is shown in green color.}
  \label{fig_compare_seg}
\end{figure}

\paragraph{Design choices of part segmentation network}
We compare our segmentation network to a version that removes the motion encoder. Additionally, we design a baseline segmentation network denoted as "Seg-baseline", which encodes the feature from a pair of input shapes independently, instead of concatenating them together, the network structure of the baseline segmentation network is provided in the supplementary. Fig.~\ref{fig_compare_seg} shows the visual results of the three methods, for each pair of inputs, we show the segmentation on the initial and end state of the point cloud. We can see that by passing the motion information to the network, the segmentation accuracy especially for the initial state is improved with fewer outliers, and the segmented region is more regular in our method compared to the other two options.
From the quantitative comparison shown in Table~\ref{tab_compare_seg}, our method achieves the best performance for the segmentation on both the start and end states of the objects.

\section{Conclusion}
\label{sec:future}

We introduce a fully automatic, active 3D reconstruction method focusing on robot-object interactions to not only recover the otherwise occluded interior structures of the target object but also obtain an understanding of part articulations.
Our approach is designed to be fully realizable by a real robot, with RGBD sensors, in a physical simulation environment. 
Qualitative and quantitative evaluations are provided to demonstrate the overall performance of our method over several object categories and articulation types.

\begin{figure}[!t]
  \includegraphics[width=0.98\linewidth]{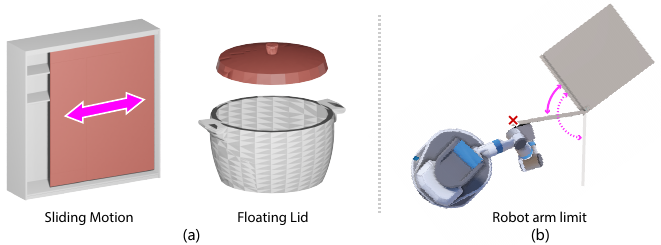}
  \caption{Failure cases of our method. (a) Two kinds of objects that our method fails to reconstruct due to the unsupported movement of the gripper or freeform motion of the non-articulated part; (b) A top-down view of a robot opening a door where the opening angle fails to reach its maximum range due to the arm limit.}
  \label{fig_failure_case}
\end{figure}

\paragraph{Limitations}
As shown in Fig.~\ref{fig_failure_case} (a) for the cabinet with a sliding motion, while the predicted action direction is correct, the suction-based gripper is unable to move in parallel to the door, leading to an unsuccessful reconstruction.
For objects with non-articulated parts such as the teapot, we can predict interactions with the lid correctly, but once the gripper picks up the lid, it is hard to continue to the next action since the gripper has to hold the lid.
Moreover, our current estimation of the part movement range may be limited by the length of the robot arm, as shown in Fig.~\ref{fig_failure_case} (b).
To initiate our interaction analysis, we require that the target 3D object to be in its rest pose, e.g., without exhibiting any part articulations. 
Similarly, both the prediction and reconstruction accuracies will hinge on proper part manipulations by the robots, e.g., complete closing of any opened parts. In reality however, error accumulation due to imprecise robot operations may be unavoidable.

In addition, our current method does not utilize any RGB information from the robot's camera, which could benefit interaction prediction.
The reconstruction quality is also limited by the sparsity of the input points. Typically, high-quality 3D reconstruction requires hundreds of thousands input points, while the raw input scans must go through a consolidation step for denoising, resampling, etc. In contrast, our problem setting involves low-end robot scanning, resulting in low scan resolution, with the timing also of concern for active 3D reconstruction with real-world interactions.

In general, each module in our method, including motion prediction, interaction execution, and mesh reconstruction, is a challenging technical problem on its own, where our current solutions all have their intrinsic limitations. Our key contribution is to demonstrate the feasibility of fully automated, robot-assisted, and interaction-driven 3D reconstruction, which should stimulate future improvements in every phase of our solution pipeline.

Our method is able to reconstruct the object interiors that are exposed by robot-object interactions, not only the part structures, but also any objects contained therein, such as a notebook inside of an opened drawer. However, for most of the shapes in synthetic dataset including ShapeNet, PartNet, the space inside are empty. A dataset with various types of objects inside is worth developing for further research.
Another future direction is to explore ways to handle hierarchical interaction, e.g., a drawer inside a cabinet door, by optimizing the interaction-perception procedure.

\section*{Acknowledgments}
We thank the reviewers for their constructive comments. This work was supported in parts by NSFC (U21B2023, U2001206, 62161146005, 62322207), DEGP Innovation Team (2022KCXTD025), GD Natural Science Foundation (2021B1515020085), Shenzhen Science and Technology Program (KQTD20210811090044003, RCJC20200714114435012, JCYJ20210324120213036), NSERC (611370), and Guangdong Laboratory of Artificial Intelligence and Digital Economy (SZ).

\bibliographystyle{ACM-Reference-Format}
\bibliography{IR_ref}


\begin{thebibliography}{38}


\ifx \showCODEN    \undefined \def \showCODEN     #1{\unskip}     \fi
\ifx \showDOI      \undefined \def \showDOI       #1{#1}\fi
\ifx \showISBNx    \undefined \def \showISBNx     #1{\unskip}     \fi
\ifx \showISBNxiii \undefined \def \showISBNxiii  #1{\unskip}     \fi
\ifx \showISSN     \undefined \def \showISSN      #1{\unskip}     \fi
\ifx \showLCCN     \undefined \def \showLCCN      #1{\unskip}     \fi
\ifx \shownote     \undefined \def \shownote      #1{#1}          \fi
\ifx \showarticletitle \undefined \def \showarticletitle #1{#1}   \fi
\ifx \showURL      \undefined \def \showURL       {\relax}        \fi
\providecommand\bibfield[2]{#2}
\providecommand\bibinfo[2]{#2}
\providecommand\natexlab[1]{#1}
\providecommand\showeprint[2][]{arXiv:#2}

\bibitem[Aloimonos et~al\mbox{.}(1988)]%
        {aloimonos1998active}
\bibfield{author}{\bibinfo{person}{J. Aloimonos}, \bibinfo{person}{I. Weiss},
  {and} \bibinfo{person}{A. Bandyopadhyay}.} \bibinfo{year}{1988}\natexlab{}.
\newblock \showarticletitle{Active vision}.
\newblock \bibinfo{journal}{\emph{Int. J. Computer Vision}}
  \bibinfo{volume}{1}, \bibinfo{number}{4} (\bibinfo{year}{1988}),
  \bibinfo{pages}{333--356}.
\newblock


\bibitem[Bajcsy(1988)]%
        {bajcsy1988active}
\bibfield{author}{\bibinfo{person}{R. Bajcsy}.}
  \bibinfo{year}{1988}\natexlab{}.
\newblock \showarticletitle{Active perception}.
\newblock \bibinfo{journal}{\emph{Proc. of the IEEE}} \bibinfo{volume}{76},
  \bibinfo{number}{8} (\bibinfo{year}{1988}), \bibinfo{pages}{966--1005}.
\newblock


\bibitem[Berger et~al\mbox{.}(2017)]%
        {berger2017survey}
\bibfield{author}{\bibinfo{person}{Matthew Berger}, \bibinfo{person}{Andrea
  Tagliasacchi}, \bibinfo{person}{Lee~M. Seversky}, \bibinfo{person}{Pierre
  Alliez}, \bibinfo{person}{Gael Guennebaud}, \bibinfo{person}{Joshua~A.
  Levine}, \bibinfo{person}{Andrei Sharf}, {and} \bibinfo{person}{Claudio~T.
  Silva}.} \bibinfo{year}{2017}\natexlab{}.
\newblock \showarticletitle{A Survey of Surface Reconstruction from Point
  Clouds}.
\newblock \bibinfo{journal}{\emph{Computer Graphics Forum}}
  \bibinfo{volume}{36} (\bibinfo{year}{2017}), \bibinfo{pages}{301--–329}.
\newblock
Issue 1.


\bibitem[Chang et~al\mbox{.}(2015)]%
        {chang2015shapenet}
\bibfield{author}{\bibinfo{person}{Angel~X. Chang}, \bibinfo{person}{Thomas
  Funkhouser}, \bibinfo{person}{Leonidas Guibas}, \bibinfo{person}{Pat
  Hanrahan}, \bibinfo{person}{Qixing Huang}, \bibinfo{person}{Zimo Li},
  \bibinfo{person}{Silvio Savarese}, \bibinfo{person}{Manolis Savva},
  \bibinfo{person}{Shuran Song}, \bibinfo{person}{Hao Su},
  \bibinfo{person}{Jianxiong Xiao}, \bibinfo{person}{Li Yi}, {and}
  \bibinfo{person}{Fisher Yu}.} \bibinfo{year}{2015}\natexlab{}.
\newblock \showarticletitle{{{ShapeNet}}: {{An Information-Rich 3D Model
  Repository}}}.
\newblock \bibinfo{journal}{\emph{arXiv:1512.03012 [cs]}}
  (\bibinfo{year}{2015}).
\newblock


\bibitem[Chen et~al\mbox{.}(2022)]%
        {chen2022ndc}
\bibfield{author}{\bibinfo{person}{Zhiqin Chen}, \bibinfo{person}{Andrea
  Tagliasacchi}, \bibinfo{person}{Thomas Funkhouser}, {and}
  \bibinfo{person}{Hao Zhang}.} \bibinfo{year}{2022}\natexlab{}.
\newblock \showarticletitle{Neural {{Dual Contouring}}}.
\newblock \bibinfo{journal}{\emph{ACM Trans. on Graphics (Proc. SIGGRAPH)}}
  \bibinfo{volume}{41}, \bibinfo{number}{4} (\bibinfo{year}{2022}),
  \bibinfo{pages}{104:1--104:13}.
\newblock


\bibitem[Chen and Zhang(2019)]%
        {chen2019imnet}
\bibfield{author}{\bibinfo{person}{Zhiqin Chen} {and} \bibinfo{person}{Hao
  Zhang}.} \bibinfo{year}{2019}\natexlab{}.
\newblock \showarticletitle{Learning implicit fields for generative shape
  modeling}. \bibinfo{pages}{5939--5948}.
\newblock


\bibitem[Gibson(1966)]%
        {gibson1966}
\bibfield{author}{\bibinfo{person}{James~J. Gibson}.}
  \bibinfo{year}{1966}\natexlab{}.
\newblock \bibinfo{booktitle}{\emph{The senses considered as perceptual
  systems}}.
\newblock \bibinfo{publisher}{Oxford England: Houghton Mifflin}.
\newblock


\bibitem[Hsu et~al\mbox{.}(2023)]%
        {hsu2023Dittoith}
\bibfield{author}{\bibinfo{person}{Cheng-Chun Hsu}, \bibinfo{person}{Zhenyu
  Jiang}, {and} \bibinfo{person}{Yuke Zhu}.} \bibinfo{year}{2023}\natexlab{}.
\newblock \showarticletitle{Ditto in the House: Building Articulation Models of
  Indoor Scenes through Interactive Perception}. In
  \bibinfo{booktitle}{\emph{Proc. IEEE Int. Conf. on Robotics \& Automation}}.
\newblock


\bibitem[Jiang et~al\mbox{.}(2022)]%
        {jiang2022ditto}
\bibfield{author}{\bibinfo{person}{Zhenyu Jiang}, \bibinfo{person}{Cheng-Chun
  Hsu}, {and} \bibinfo{person}{Yuke Zhu}.} \bibinfo{year}{2022}\natexlab{}.
\newblock \showarticletitle{Ditto: {{Building Digital Twins}} of {{Articulated
  Objects}} from {{Interaction}}}. In \bibinfo{booktitle}{\emph{Proc. IEEE/CVF
  Conf. on Computer Vision \& Pattern Recognition}}.
  \bibinfo{pages}{5616--5626}.
\newblock


\bibitem[Katz et~al\mbox{.}(2013)]%
        {katz2013interactive}
\bibfield{author}{\bibinfo{person}{Dov Katz}, \bibinfo{person}{Moslem Kazemi},
  \bibinfo{person}{J. Andrew~Bagnell}, {and} \bibinfo{person}{Anthony Stentz}.}
  \bibinfo{year}{2013}\natexlab{}.
\newblock \showarticletitle{Interactive Segmentation, Tracking, and Kinematic
  Modeling of Unknown {{3D}} Articulated Objects}. In
  \bibinfo{booktitle}{\emph{Proc. IEEE Int. Conf. on Robotics \& Automation}}.
  \bibinfo{pages}{5003--5010}.
\newblock


\bibitem[Kawana et~al\mbox{.}(2022)]%
        {kawana2022unsupervised}
\bibfield{author}{\bibinfo{person}{Yuki Kawana}, \bibinfo{person}{Yusuke
  Mukuta}, {and} \bibinfo{person}{Tatsuya Harada}.}
  \bibinfo{year}{2022}\natexlab{}.
\newblock \showarticletitle{Unsupervised {{Pose-aware Part Decomposition}} for
  {{Man-Made Articulated Objects}}}. In \bibinfo{booktitle}{\emph{Proc. Euro.
  Conf. on Computer Vision}}. \bibinfo{pages}{558--575}.
\newblock


\bibitem[Li et~al\mbox{.}(2022)]%
        {li2022igibson}
\bibfield{author}{\bibinfo{person}{Chengshu Li}, \bibinfo{person}{Fei Xia},
  \bibinfo{person}{Roberto {Mart{\'i}n-Mart{\'i}n}}, \bibinfo{person}{Michael
  Lingelbach}, \bibinfo{person}{Sanjana Srivastava}, \bibinfo{person}{Bokui
  Shen}, \bibinfo{person}{Kent~Elliott Vainio}, \bibinfo{person}{Cem Gokmen},
  \bibinfo{person}{Gokul Dharan}, \bibinfo{person}{Tanish Jain},
  \bibinfo{person}{Andrey Kurenkov}, \bibinfo{person}{Karen Liu},
  \bibinfo{person}{Hyowon Gweon}, \bibinfo{person}{Jiajun Wu},
  \bibinfo{person}{Li {Fei-Fei}}, {and} \bibinfo{person}{Silvio Savarese}.}
  \bibinfo{year}{2022}\natexlab{}.
\newblock \showarticletitle{{{iGibson}} 2.0: {{Object-Centric Simulation}} for
  {{Robot Learning}} of {{Everyday Household Tasks}}}. In
  \bibinfo{booktitle}{\emph{Proc. Conf. on Robot Learning}},
  Vol.~\bibinfo{volume}{164}. \bibinfo{pages}{455--465}.
\newblock


\bibitem[Liu et~al\mbox{.}(2022)]%
        {liu2022akb48}
\bibfield{author}{\bibinfo{person}{Liu Liu}, \bibinfo{person}{Wenqiang Xu},
  \bibinfo{person}{Haoyuan Fu}, \bibinfo{person}{Sucheng Qian},
  \bibinfo{person}{Qiaojun Yu}, \bibinfo{person}{Yang Han}, {and}
  \bibinfo{person}{Cewu Lu}.} \bibinfo{year}{2022}\natexlab{}.
\newblock \showarticletitle{{{AKB-48}}: {{A Real-World Articulated Object
  Knowledge Base}}}. In \bibinfo{booktitle}{\emph{Proc. IEEE/CVF Conf. on
  Computer Vision \& Pattern Recognition}}. \bibinfo{pages}{14809--14818}.
\newblock


\bibitem[{Mart{\'i}n-Mart{\'i}n} et~al\mbox{.}(2016)]%
        {martin2016integrated}
\bibfield{author}{\bibinfo{person}{Roberto {Mart{\'i}n-Mart{\'i}n}},
  \bibinfo{person}{Sebastian H{\"o}fer}, {and} \bibinfo{person}{Oliver Brock}.}
  \bibinfo{year}{2016}\natexlab{}.
\newblock \showarticletitle{An {{Integrated Approach}} to {{Visual Perception}}
  of {{Articulated Objects}}}. In \bibinfo{booktitle}{\emph{Proc. IEEE Int.
  Conf. on Robotics \& Automation}}. \bibinfo{pages}{5091--5097}.
\newblock


\bibitem[Mescheder et~al\mbox{.}(2019)]%
        {mescheder2019occupancy}
\bibfield{author}{\bibinfo{person}{Lars Mescheder}, \bibinfo{person}{Michael
  Oechsle}, \bibinfo{person}{Michael Niemeyer}, \bibinfo{person}{Sebastian
  Nowozin}, {and} \bibinfo{person}{Andreas Geiger}.}
  \bibinfo{year}{2019}\natexlab{}.
\newblock \showarticletitle{Occupancy {{Networks}}: {{Learning 3D
  Reconstruction}} in {{Function Space}}}. In \bibinfo{booktitle}{\emph{Proc.
  IEEE/CVF Conf. on Computer Vision \& Pattern Recognition}}.
  \bibinfo{pages}{4455--4465}.
\newblock


\bibitem[Mo et~al\mbox{.}(2021)]%
        {mo2021where2act}
\bibfield{author}{\bibinfo{person}{Kaichun Mo}, \bibinfo{person}{Leonidas
  Guibas}, \bibinfo{person}{Mustafa Mukadam}, \bibinfo{person}{Abhinav Gupta},
  {and} \bibinfo{person}{Shubham Tulsiani}.} \bibinfo{year}{2021}\natexlab{}.
\newblock \showarticletitle{{{Where2Act}}: {{From Pixels}} to {{Actions}} for
  {{Articulated 3D Objects}}}. In \bibinfo{booktitle}{\emph{Proc. Int. Conf. on
  Computer Vision}}. \bibinfo{pages}{6793--6803}.
\newblock


\bibitem[Mo et~al\mbox{.}(2019)]%
        {mo2019partnet}
\bibfield{author}{\bibinfo{person}{Kaichun Mo}, \bibinfo{person}{Shilin Zhu},
  \bibinfo{person}{Angel~X. Chang}, \bibinfo{person}{Li Yi},
  \bibinfo{person}{Subarna Tripathi}, \bibinfo{person}{Leonidas~J. Guibas},
  {and} \bibinfo{person}{Hao Su}.} \bibinfo{year}{2019}\natexlab{}.
\newblock \showarticletitle{{{PartNet}}: {{A Large-Scale Benchmark}} for
  {{Fine-Grained}} and {{Hierarchical Part-Level 3D Object Understanding}}}. In
  \bibinfo{booktitle}{\emph{Proc. IEEE/CVF Conf. on Computer Vision \& Pattern
  Recognition}}. \bibinfo{pages}{909--918}.
\newblock


\bibitem[Mu et~al\mbox{.}(2021)]%
        {mu2021asdf}
\bibfield{author}{\bibinfo{person}{Jiteng Mu}, \bibinfo{person}{Weichao Qiu},
  \bibinfo{person}{Adam Kortylewski}, \bibinfo{person}{Alan Yuille},
  \bibinfo{person}{Nuno Vasconcelos}, {and} \bibinfo{person}{Xiaolong Wang}.}
  \bibinfo{year}{2021}\natexlab{}.
\newblock \showarticletitle{A-{{SDF}}: {{Learning Disentangled Signed Distance
  Functions}} for {{Articulated Shape Representation}}}. In
  \bibinfo{booktitle}{\emph{Proc. Int. Conf. on Computer Vision}}.
  \bibinfo{pages}{12981--12991}.
\newblock


\bibitem[Newcombe et~al\mbox{.}(2011)]%
        {newcombe2011kinect}
\bibfield{author}{\bibinfo{person}{Richard~A. Newcombe},
  \bibinfo{person}{Shahram Izadi}, \bibinfo{person}{Otmar Hilliges},
  \bibinfo{person}{David Molyneaux}, \bibinfo{person}{David Kim},
  \bibinfo{person}{Andrew~J. Davison}, \bibinfo{person}{Pushmeet Kohi},
  \bibinfo{person}{Jamie Shotton}, \bibinfo{person}{Steve Hodges}, {and}
  \bibinfo{person}{Andrew Fitzgibbon}.} \bibinfo{year}{2011}\natexlab{}.
\newblock \showarticletitle{{{KinectFusion}}: {{Real-time}} Dense Surface
  Mapping and Tracking}. In \bibinfo{booktitle}{\emph{Proc. IEEE Int. Symp. on
  Mixed and augmented reality}}. \bibinfo{pages}{127--136}.
\newblock


\bibitem[Nie et~al\mbox{.}(2023)]%
        {nie2023structure}
\bibfield{author}{\bibinfo{person}{Neil Nie}, \bibinfo{person}{Samir~Yitzhak
  Gadre}, \bibinfo{person}{Kiana Ehsani}, {and} \bibinfo{person}{Shuran Song}.}
  \bibinfo{year}{2023}\natexlab{}.
\newblock \bibinfo{title}{Structure from {{Action}}: {{Learning Interactions}}
  for {{Articulated Object 3D Structure Discovery}}}.
\newblock
\newblock


\bibitem[Paolillo et~al\mbox{.}(2017)]%
        {paolillo2017visual}
\bibfield{author}{\bibinfo{person}{Antonio Paolillo},
  \bibinfo{person}{Anastasia Bolotnikova}, \bibinfo{person}{K{\'e}vin
  Chappellet}, {and} \bibinfo{person}{Abderrahmane Kheddar}.}
  \bibinfo{year}{2017}\natexlab{}.
\newblock \showarticletitle{Visual Estimation of Articulated Objects
  Configuration during Manipulation with a Humanoid}. In
  \bibinfo{booktitle}{\emph{Proc. IEEE/SICE Int. Symp. on System Integration}}.
  \bibinfo{pages}{330--335}.
\newblock


\bibitem[Park et~al\mbox{.}(2019)]%
        {park2019deepsdf}
\bibfield{author}{\bibinfo{person}{Jeong~Joon Park}, \bibinfo{person}{Peter
  Florence}, \bibinfo{person}{Julian Straub}, \bibinfo{person}{Richard
  Newcombe}, {and} \bibinfo{person}{Steven Lovegrove}.}
  \bibinfo{year}{2019}\natexlab{}.
\newblock \showarticletitle{{DeepSDF}: Learning continuous signed distance
  functions for shape representation}. In \bibinfo{booktitle}{\emph{Proc.
  IEEE/CVF Conf. on Computer Vision \& Pattern Recognition}}.
  \bibinfo{pages}{165--174}.
\newblock


\bibitem[Puig et~al\mbox{.}(2018)]%
        {puig2018virtualhome}
\bibfield{author}{\bibinfo{person}{Xavier Puig}, \bibinfo{person}{Kevin Ra},
  \bibinfo{person}{Marko Boben}, \bibinfo{person}{Jiaman Li},
  \bibinfo{person}{Tingwu Wang}, \bibinfo{person}{Sanja Fidler}, {and}
  \bibinfo{person}{Antonio Torralba}.} \bibinfo{year}{2018}\natexlab{}.
\newblock \showarticletitle{{{VirtualHome}}: {{Simulating Household
  Activities}} via {{Programs}}}. In \bibinfo{booktitle}{\emph{Proc. IEEE/CVF
  Conf. on Computer Vision \& Pattern Recognition}}.
  \bibinfo{pages}{8494--8502}.
\newblock


\bibitem[Qian et~al\mbox{.}(2022)]%
        {qian2022pointnext}
\bibfield{author}{\bibinfo{person}{Guocheng Qian}, \bibinfo{person}{Yuchen Li},
  \bibinfo{person}{Houwen Peng}, \bibinfo{person}{Jinjie Mai},
  \bibinfo{person}{Hasan Hammoud}, \bibinfo{person}{Mohamed Elhoseiny}, {and}
  \bibinfo{person}{Bernard Ghanem}.} \bibinfo{year}{2022}\natexlab{}.
\newblock \showarticletitle{PointNeXt: Revisiting PointNet++ with Improved
  Training and Scaling Strategies}. In \bibinfo{booktitle}{\emph{Proc. Conf. on
  Neural Information Processing Systems}}.
\newblock


\bibitem[Shen et~al\mbox{.}(2021)]%
        {shen2021igibson}
\bibfield{author}{\bibinfo{person}{Bokui Shen}, \bibinfo{person}{Fei Xia},
  \bibinfo{person}{Chengshu Li}, \bibinfo{person}{Roberto
  {Mart{\'i}n-Mart{\'i}n}}, \bibinfo{person}{Linxi Fan},
  \bibinfo{person}{Guanzhi Wang}, \bibinfo{person}{Claudia
  {P{\'e}rez-D'Arpino}}, \bibinfo{person}{Shyamal Buch},
  \bibinfo{person}{Sanjana Srivastava}, \bibinfo{person}{Lyne Tchapmi},
  \bibinfo{person}{Micael Tchapmi}, \bibinfo{person}{Kent Vainio},
  \bibinfo{person}{Josiah Wong}, \bibinfo{person}{Li {Fei-Fei}}, {and}
  \bibinfo{person}{Silvio Savarese}.} \bibinfo{year}{2021}\natexlab{}.
\newblock \showarticletitle{{{iGibson}} 1.0: {{A Simulation Environment}} for
  {{Interactive Tasks}} in {{Large Realistic Scenes}}}. In
  \bibinfo{booktitle}{\emph{Proc. IEEE Int. Conf. on Intelligent Robots \&
  Systems}}. \bibinfo{pages}{7520--7527}.
\newblock


\bibitem[Wang et~al\mbox{.}(2019)]%
        {wang2019shape2motion}
\bibfield{author}{\bibinfo{person}{Xiaogang Wang}, \bibinfo{person}{Bin Zhou},
  \bibinfo{person}{Yahao Shi}, \bibinfo{person}{Xiaowu Chen},
  \bibinfo{person}{Qinping Zhao}, {and} \bibinfo{person}{Kai Xu}.}
  \bibinfo{year}{2019}\natexlab{}.
\newblock \showarticletitle{{{Shape2Motion}}: {{Joint Analysis}} of {{Motion
  Parts}} and {{Attributes From 3D Shapes}}}. In
  \bibinfo{booktitle}{\emph{Proc. IEEE/CVF Conf. on Computer Vision \& Pattern
  Recognition}}. \bibinfo{pages}{8868--8876}.
\newblock


\bibitem[Wang et~al\mbox{.}(2022)]%
        {wang2022adaafford}
\bibfield{author}{\bibinfo{person}{Yian Wang}, \bibinfo{person}{Ruihai Wu},
  \bibinfo{person}{Kaichun Mo}, \bibinfo{person}{Jiaqi Ke},
  \bibinfo{person}{Qingnan Fan}, \bibinfo{person}{Leonidas~J. Guibas}, {and}
  \bibinfo{person}{Hao Dong}.} \bibinfo{year}{2022}\natexlab{}.
\newblock \showarticletitle{{{AdaAfford}}: {{Learning}} to~{{Adapt Manipulation
  Affordance}} for~{{3D Articulated Objects}} via~{{Few-Shot Interactions}}}.
  In \bibinfo{booktitle}{\emph{Proc. Euro. Conf. on Computer Vision}}.
  \bibinfo{pages}{90--107}.
\newblock


\bibitem[Wei et~al\mbox{.}(2022)]%
        {wei2022self}
\bibfield{author}{\bibinfo{person}{Fangyin Wei}, \bibinfo{person}{Rohan
  Chabra}, \bibinfo{person}{Lingni Ma}, \bibinfo{person}{Christoph Lassner},
  \bibinfo{person}{Michael Zollh{\"o}fer}, \bibinfo{person}{Szymon
  Rusinkiewicz}, \bibinfo{person}{Chris Sweeney}, \bibinfo{person}{Richard
  Newcombe}, {and} \bibinfo{person}{Mira Slavcheva}.}
  \bibinfo{year}{2022}\natexlab{}.
\newblock \showarticletitle{Self-supervised Neural Articulated Shape and
  Appearance Models}. In \bibinfo{booktitle}{\emph{Proc. IEEE/CVF Conf. on
  Computer Vision \& Pattern Recognition}}. \bibinfo{pages}{15816--15826}.
\newblock


\bibitem[Wu et~al\mbox{.}(2014)]%
        {wu2014qualitydriven}
\bibfield{author}{\bibinfo{person}{Shihao Wu}, \bibinfo{person}{Wei Sun},
  \bibinfo{person}{Pinxin Long}, \bibinfo{person}{Hui Huang},
  \bibinfo{person}{Daniel {Cohen-Or}}, \bibinfo{person}{Minglun Gong},
  \bibinfo{person}{Oliver Deussen}, {and} \bibinfo{person}{Baoquan Chen}.}
  \bibinfo{year}{2014}\natexlab{}.
\newblock \showarticletitle{Quality-Driven {{Poisson-guided Autoscanning}}}.
\newblock \bibinfo{journal}{\emph{ACM Trans. on Graphics (Proc. SIGGRAPH
  Asia)}} \bibinfo{volume}{33}, \bibinfo{number}{6} (\bibinfo{year}{2014}),
  \bibinfo{pages}{203:1--203:12}.
\newblock


\bibitem[Xiang et~al\mbox{.}(2020)]%
        {xiang2020sapien}
\bibfield{author}{\bibinfo{person}{Fanbo Xiang}, \bibinfo{person}{Yuzhe Qin},
  \bibinfo{person}{Kaichun Mo}, \bibinfo{person}{Yikuan Xia},
  \bibinfo{person}{Hao Zhu}, \bibinfo{person}{Fangchen Liu},
  \bibinfo{person}{Minghua Liu}, \bibinfo{person}{Hanxiao Jiang},
  \bibinfo{person}{Yifu Yuan}, \bibinfo{person}{He Wang}, \bibinfo{person}{Li
  Yi}, \bibinfo{person}{Angel~X. Chang}, \bibinfo{person}{Leonidas~J. Guibas},
  {and} \bibinfo{person}{Hao Su}.} \bibinfo{year}{2020}\natexlab{}.
\newblock \showarticletitle{{{SAPIEN}}: {{A SimulAted Part-based Interactive
  ENvironment}}}. In \bibinfo{booktitle}{\emph{Proc. IEEE/CVF Conf. on Computer
  Vision \& Pattern Recognition}}. \bibinfo{pages}{11094--11104}.
\newblock


\bibitem[Xie et~al\mbox{.}(2022)]%
        {NF_survey}
\bibfield{author}{\bibinfo{person}{Yiheng Xie}, \bibinfo{person}{Towaki
  Takikawa}, \bibinfo{person}{Shunsuke Saito}, \bibinfo{person}{Or Litany},
  \bibinfo{person}{Shiqin Yan}, \bibinfo{person}{Numair Khan},
  \bibinfo{person}{Federico Tombari}, \bibinfo{person}{James Tompkin},
  \bibinfo{person}{Vincent Sitzmann}, {and} \bibinfo{person}{Srinath Sridhar}.}
  \bibinfo{year}{2022}\natexlab{}.
\newblock \showarticletitle{Neural Fields in Visual Computing and Beyond}.
\newblock \bibinfo{journal}{\emph{Computer Graphics Forum}}
  (\bibinfo{year}{2022}).
\newblock


\bibitem[Yan et~al\mbox{.}(2014)]%
        {yan2014proactive}
\bibfield{author}{\bibinfo{person}{Feilong Yan}, \bibinfo{person}{Andrei
  Sharf}, \bibinfo{person}{Wenzhen Lin}, \bibinfo{person}{Hui Huang}, {and}
  \bibinfo{person}{Baoquan Chen}.} \bibinfo{year}{2014}\natexlab{}.
\newblock \showarticletitle{Proactive {{3D Scanning}} of {{Inaccessible
  Parts}}}.
\newblock \bibinfo{journal}{\emph{ACM Trans. on Graphics (Proc. SIGGRAPH)}}
  \bibinfo{volume}{33}, \bibinfo{number}{4} (\bibinfo{year}{2014}),
  \bibinfo{pages}{1--8}.
\newblock


\bibitem[Yan and Pollefeys(2006)]%
        {yan2006automatic}
\bibfield{author}{\bibinfo{person}{Jingyu Yan} {and} \bibinfo{person}{Marc
  Pollefeys}.} \bibinfo{year}{2006}\natexlab{}.
\newblock \showarticletitle{Automatic {{Kinematic Chain Building}} from
  {{Feature Trajectories}} of {{Articulated Objects}}}. In
  \bibinfo{booktitle}{\emph{Proc. IEEE/CVF Conf. on Computer Vision \& Pattern
  Recognition}}. \bibinfo{pages}{712--719}.
\newblock


\bibitem[Yan et~al\mbox{.}(2019)]%
        {yan2019rpmnet}
\bibfield{author}{\bibinfo{person}{Zihao Yan}, \bibinfo{person}{Ruizhen Hu},
  \bibinfo{person}{Xingguang Yan}, \bibinfo{person}{Luanmin Chen},
  \bibinfo{person}{Oliver Van~Kaick}, \bibinfo{person}{Hao Zhang}, {and}
  \bibinfo{person}{Hui Huang}.} \bibinfo{year}{2019}\natexlab{}.
\newblock \showarticletitle{{{RPM-Net}}: {{Recurrent Prediction}} of {{Motion}}
  and {{Parts From Point Cloud}}}.
\newblock \bibinfo{journal}{\emph{ACM Trans. on Graphics (Proc. SIGGRAPH
  Asia)}} \bibinfo{volume}{38}, \bibinfo{number}{6} (\bibinfo{year}{2019}),
  \bibinfo{pages}{1--15}.
\newblock


\bibitem[Yang et~al\mbox{.}(2018)]%
        {yang2018active}
\bibfield{author}{\bibinfo{person}{Xin Yang}, \bibinfo{person}{Yuanbo Wang},
  \bibinfo{person}{Yaru Wang}, \bibinfo{person}{Baocai Yin},
  \bibinfo{person}{Qiang Zhang}, \bibinfo{person}{Xiaopeng Wei}, {and}
  \bibinfo{person}{Hongbo Fu}.} \bibinfo{year}{2018}\natexlab{}.
\newblock \showarticletitle{Active Object Reconstruction Using a Guided View
  Planner}. In \bibinfo{booktitle}{\emph{Proc. Int. Joint Conf. on Artificial
  Intelligence}}. \bibinfo{pages}{4965--4971}.
\newblock


\bibitem[Yi et~al\mbox{.}(2018)]%
        {yi2018deep}
\bibfield{author}{\bibinfo{person}{Li Yi}, \bibinfo{person}{Haibin Huang},
  \bibinfo{person}{Difan Liu}, \bibinfo{person}{Evangelos Kalogerakis},
  \bibinfo{person}{Hao Su}, {and} \bibinfo{person}{Leonidas Guibas}.}
  \bibinfo{year}{2018}\natexlab{}.
\newblock \showarticletitle{Deep {{Part Induction}} from {{Articulated Object
  Pairs}}}.
\newblock \bibinfo{journal}{\emph{ACM Trans. on Graphics (Proc. SIGGRAPH
  Asia)}}  \bibinfo{volume}{37} (\bibinfo{year}{2018}),
  \bibinfo{pages}{209:1--209:15}.
\newblock


\bibitem[Zhang et~al\mbox{.}(2021)]%
        {zhang2021strobenet}
\bibfield{author}{\bibinfo{person}{Ge Zhang}, \bibinfo{person}{Or Litany},
  \bibinfo{person}{Srinath Sridhar}, {and} \bibinfo{person}{Leonidas Guibas}.}
  \bibinfo{year}{2021}\natexlab{}.
\newblock \bibinfo{title}{{{StrobeNet}}: {{Category-Level Multiview
  Reconstruction}} of {{Articulated Objects}}}.
\newblock
\newblock


\bibitem[Zhou et~al\mbox{.}(2022)]%
        {zhou2022seedformer}
\bibfield{author}{\bibinfo{person}{Haoran Zhou}, \bibinfo{person}{Yun Cao},
  \bibinfo{person}{Wenqing Chu}, \bibinfo{person}{Junwei Zhu},
  \bibinfo{person}{Tong Lu}, \bibinfo{person}{Ying Tai}, {and}
  \bibinfo{person}{Chengjie Wang}.} \bibinfo{year}{2022}\natexlab{}.
\newblock \showarticletitle{{{SeedFormer}}: {{Patch Seeds Based Point Cloud
  Completion}} with~{{Upsample Transformer}}}. In
  \bibinfo{booktitle}{\emph{Proc. Euro. Conf. on Computer Vision}}.
  \bibinfo{pages}{416--432}.
\newblock


\end{thebibliography}

\end{document}